\documentclass[11pt,a4paper]{article}
\usepackage{times}
\usepackage[T2A,T1]{fontenc}
\usepackage{latexsym}
\usepackage{url}

\usepackage{booktabs}
\usepackage{multirow}
\usepackage{graphicx}

\usepackage[dvipsnames]{xcolor}
\usepackage{microtype}
\usepackage{enumitem} 
\usepackage{tikz}

\usepackage{pifont} 
\usepackage{pgf}
\usepackage{colortbl}

\DeclareRobustCommand{\cyr}[1]{%
  {\fontencoding{T2A}\fontfamily{ftm}\selectfont #1}%
}

\newcommand{\gradientcell}[6]{%
    \def\value{#1}%
    \def\minvalue{#2}%
    \def\maxvalue{#3}%
    \def\mincolor{#4}%
    \def\maxcolor{#5}%
    \def\transparency{#6}%
    \ifdimcomp{\value pt}{>}{\maxvalue pt}{\cellcolor{#5!100.0!#4!#6}\value}{%
    \ifdimcomp{\value pt}{<}{\minvalue pt}{\cellcolor{#5!0.0!#4!#6}\value}{%
         \pgfmathparse{int(round(100*(#1/(\maxvalue-\minvalue))-(\minvalue *(100/(\maxvalue-\minvalue)))))}%
        \xdef\tempa{\pgfmathresult}%
        \cellcolor{#5!\tempa!#4!#6}\value%
    }}%
}

\definecolor{LightGray}{gray}{0.95}
\newcommand{\Cat}[1]{\gradientcell{#1}{0}{60}{LightGray}{Green}{60}}
\newcommand{\Ber}[1]{\gradientcell{#1}{10}{40}{LightGray}{Bittersweet}{60}}
\newcommand{\Ner}[1]{\gradientcell{#1}{0}{90}{LightGray}{Goldenrod}{60}}
\newcommand{\agreement}[1]{\gradientcell{#1}{0.4672566371681417}{0.9274509803921569}{LightGray}{Goldenrod}{60}}

\newcommand{\XlingTextualCZ}[1]{\gradientcell{#1}{5}{49}{LightGray}{Cerulean}{60}}
\newcommand{\XlingTextualSK}[1]{\gradientcell{#1}{4}{43}{LightGray}{Bittersweet}{60}}
\newcommand{\XlingTextualUA}[1]{\gradientcell{#1}{7}{42}{LightGray}{Goldenrod}{60}}
\newcommand{\XlingVisualCZ}[1]{\gradientcell{#1}{-1}{9}{LightGray}{Cerulean}{30}}
\newcommand{\XlingVisualSK}[1]{\gradientcell{#1}{-1}{15}{LightGray}{Bittersweet}{30}}
\newcommand{\XlingVisualUA}[1]{\gradientcell{#1}{-1}{21}{LightGray}{Goldenrod}{30}}

\newcommand{\human}[1]{\gradientcell{#1}{0}{58}{LightGray}{LimeGreen}{60}}
\newcommand{\bleu}[1]{\gradientcell{#1}{0.5}{9.5}{LightGray}{Cerulean}{60}}
\newcommand{\chrf}[1]{\gradientcell{#1}{5}{42}{LightGray}{Cerulean}{60}}
\newcommand{\rouge}[1]{\gradientcell{#1}{.004}{0.364}{LightGray}{Cerulean}{60}}
\newcommand{\bertscore}[1]{\gradientcell{#1}{0}{.755}{LightGray}{Cerulean}{60}}
\newcommand{\bleurt}[1]{\gradientcell{#1}{.16}{.50}{LightGray}{Cerulean}{60}}
\newcommand{\llamascore}[1]{\gradientcell{#1}{0}{50}{LightGray}{Cerulean}{60}}
\newcommand{\llamaallscore}[1]{\gradientcell{#1}{0}{85}{LightGray}{Cerulean}{60}}
\newcommand{\prometheus}[1]{\gradientcell{#1}{0}{48}{LightGray}{Cerulean}{60}}

\newcommand{\Co}[1]{\gradientcell{#1}{.5}{1.0}{LightGray}{Bittersweet}{60}}
\newcommand{\R}[1]{\gradientcell{#1}{.3}{1.0}{LightGray}{BlueGreen}{60}}

\definecolor{tptnlow}{RGB}{235,248,235}
\definecolor{tptnhigh}{RGB}{100,180,100}
\definecolor{fpfnlow}{RGB}{255,245,245}
\definecolor{fpfnhigh}{RGB}{240,120,120}

\newcommand{\cmcellsize}{0.4cm}

\newcommand{\confmat}[4]{%
  \begin{tikzpicture}[baseline=-0.5ex]
    \pgfmathtruncatemacro{\tpmix}{100-#1}
    \pgfmathtruncatemacro{\fnmix}{100-#2}
    \pgfmathtruncatemacro{\fpmix}{100-#3}
    \pgfmathtruncatemacro{\tnmix}{100-#4}
    
    \fill[tptnlow!\tpmix!tptnhigh] (0,0) rectangle (\cmcellsize,\cmcellsize);
    \node[font=\tiny] at (0.5*\cmcellsize,0.5*\cmcellsize) {#1};
    
    \fill[fpfnlow!\fnmix!fpfnhigh] (\cmcellsize,0) rectangle (2*\cmcellsize,\cmcellsize);
    \node[font=\tiny] at (1.5*\cmcellsize,0.5*\cmcellsize) {#2};
    
    \fill[fpfnlow!\fpmix!fpfnhigh] (0,-\cmcellsize) rectangle (\cmcellsize,0);
    \node[font=\tiny] at (0.5*\cmcellsize,-0.5*\cmcellsize) {#3};
    
    \fill[tptnlow!\tnmix!tptnhigh] (\cmcellsize,-\cmcellsize) rectangle (2*\cmcellsize,0);
    \node[font=\tiny] at (1.5*\cmcellsize,-0.5*\cmcellsize) {#4};
    
    \draw[gray] (0,-\cmcellsize) rectangle (2*\cmcellsize,\cmcellsize);
    \node[font=\tiny] at (1.5*\cmcellsize,-0.5*\cmcellsize) {#4};
  \end{tikzpicture}%
}

\newcommand{\confmatlegend}{%
  \begin{tikzpicture}[baseline=-0.5ex]
    \fill[white] (0,0) rectangle (\cmcellsize,\cmcellsize);
    \node[font=\tiny] at (0.5*\cmcellsize,0.5*\cmcellsize) {TP};
    
    \fill[gray!25] (\cmcellsize,0) rectangle (2*\cmcellsize,\cmcellsize);
    \node[font=\tiny] at (1.5*\cmcellsize,0.5*\cmcellsize) {FN};
    
    \fill[gray!25] (0,-\cmcellsize) rectangle (\cmcellsize,0);
    \node[font=\tiny] at (0.5*\cmcellsize,-0.5*\cmcellsize) {FP};
    
    \fill[white] (\cmcellsize,-\cmcellsize) rectangle (2*\cmcellsize,0);
    \node[font=\tiny] at (1.5*\cmcellsize,-0.5*\cmcellsize) {TN};
    
    \draw[gray] (0,-\cmcellsize) rectangle (2*\cmcellsize,\cmcellsize);
  \end{tikzpicture}%
}

\usepackage[acceptedWithA]{tacl2021v1}

\usepackage{xspace,mfirstuc,tabulary}

\newif\iftaclinstructions
\taclinstructionsfalse 
\iftaclinstructions
\renewcommand{\confidential}{}
\renewcommand{\anonsubtext}{(No author info supplied here, for consistency with
TACL-submission anonymization requirements)}
\newcommand{\instr}
\fi

\iftaclpubformat 

\else

\fi

\title{{CUS-QA}: Local-Knowledge-Oriented Open-Ended \\ Question Answering Dataset}

\author{
    Jindřich Libovický\textsuperscript{1}\thanks{~~Equal contribution} \quad Jindřich Helcl\textsuperscript{2}$^\ast$ \quad Andrei-Alexandru Manea\textsuperscript{1} \quad Gianluca Vico\textsuperscript{1}
    \\ \\
  \textsuperscript{1}Charles University \qquad \textsuperscript{2}University of Oslo
}

\date{}

\begin{document}
\maketitle

\begin{abstract}
We introduce CUS-QA, a benchmark for evaluation of open-ended regional question answering that encompasses both textual and visual modalities. We also provide strong baselines using state-of-the-art large language models (LLMs).
Our dataset consists of manually curated questions and answers grounded in Wikipedia, created by native speakers from Czechia, Slovakia, and Ukraine, with accompanying English translations. It includes both purely textual questions and those requiring visual understanding.
We evaluate state-of-the-art LLMs through prompting and add human judgments of answer correctness. Using these human evaluations, we analyze the reliability of existing automatic evaluation metrics.
Our baseline results show that even the best open-weight LLMs achieve only over 40\% accuracy on textual questions and below 30\% on visual questions. LLM-based evaluation metrics show strong correlation with human judgment, while traditional string-overlap metrics perform surprisingly well due to the prevalence of named entities in answers.
%
%
\end{abstract}

\section{Introduction}


The ability to answer factual questions is one of the most frequently used and evaluated capabilities of current large language models (LLMs), both purely textual and multimodal. However, existing evaluation approaches have significant limitations that fail to capture realistic usage scenarios. 

Most widely used QA benchmarks focus on globally known facts and use multiple-choice formats for easy evaluation and cross-lingual comparison \citep{hendrycks2021measuring,clark-etal-2020-tydi,rajpurkar-etal-2016-squad}. This setup is unrealistic in two important ways. First, when people use LLMs to get answers to their questions, they ask open-ended questions without providing multiple answer choices. Second, the questions that matter to users vary significantly across regions and cultures. A question about local geography, politics, or cultural figures that is highly relevant to speakers of one language may be completely irrelevant to speakers of another language.

Regional knowledge presents a particularly challenging test case for LLMs. Unlike globally known facts, regional information requires models to have learned culturally specific knowledge during training, often from less-represented languages and sources.
Moreover, question-answering performance of LLMs varies wildly depending on the language of the question \citep{singh2024global}.
This makes region-specific QA an ideal probe for understanding the knowledge gaps and cultural biases in current models.

\begin{table}[t!]
\centering\footnotesize
\setlength{\tabcolsep}{3.5pt}

\begin{tabular}{l p{2.5cm} p{1.5cm}}
\toprule
Image & Question & Answer \\ \midrule

\multirow{2}{*}{\includegraphics[width=3cm]{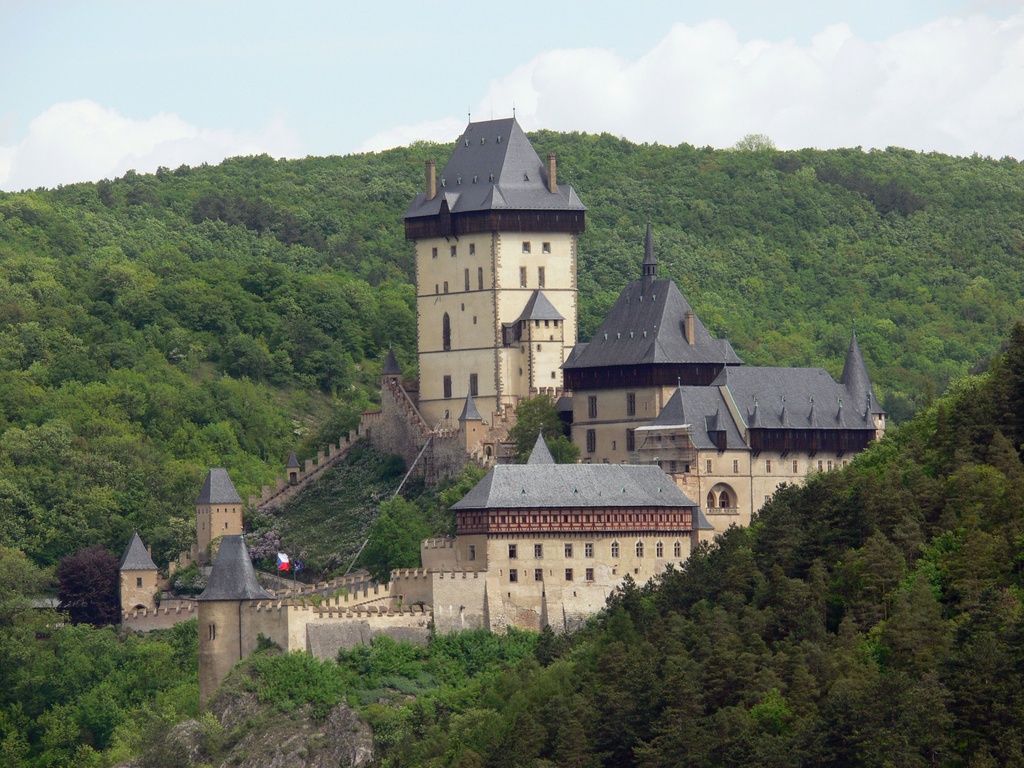}} & Kdo nechal postavit známý český hrad na obrázku? & Karel IV. \\ \addlinespace
& \it Who built the famous Czech castle in the picture? & \it Charles IV. \\[22pt]

\bottomrule
\end{tabular}

\caption{Visual sample of collected Czech questions and answers.}\label{tab:example_short}
\end{table}

Open-ended question answering poses an additional evaluation challenge beyond the regional knowledge gap. Unlike the multiple-choice setup, open-ended responses cannot be evaluated through exact matching, as there are typically multiple ways to express the correct answer. 
Furthermore, evaluation of open-ended QA systems remains an underexplored area, with
not enough resources that can be used to validate the evaluation metrics.

We introduce CUS-QA (Czech-Ukrainian-Slovak Question Answering), a dataset that addresses both limitations simultaneously. We collect questions about regional knowledge from native speakers of Czech, Slovak, and Ukrainian, three Slavic languages with varying resource levels and speaker populations. Our questions are grounded in local Wikipedias and cover both textual and visual modalities and require models to show knowledge about local geography, culture, history, and politics that is well-known within each country but largely unknown outside it.

To establish baselines and understand the challenges posed by our dataset, we evaluate several state-of-the-art open-weight LLMs and complement automatic evaluation with human judgment. Our results reveal significant gaps in regional knowledge: even the best models achieve only over 40\% accuracy on textual questions and below 30\% on visual questions. We also find substantial cross-lingual inconsistencies, where models sometimes perform better at answering regional questions in English rather than the local language (\S~\ref{ssec:xling_diff}). 

Beyond establishing performance baselines, we conduct a human evaluation to analyze the reliability of automatic evaluation metrics for open-ended QA. This addresses the critical need in the field for human-labeled datasets to develop metrics for reference-based evaluation of open-ended generation.

The paper is structured as follows: Section~\ref{sec:collection} describes our data collection process, followed by a detailed dataset description in Section~\ref{sec:dataset}. Section~\ref{sec:evaluation} presents our evaluation methodology and baseline approaches. Baseline results, correlation analysis between automatic metrics and human judgment, cross-lingual differences, analysis of robustness to prompt variations, and results of retrieval-augmented generation experiments are described in Section~\ref{sec:results}. We discuss implications and future work in Section~\ref{sec:discussion}, give an overview of related work on multilingual QA and evaluation metrics in Section~\ref{sec:related_work}, and conclude in Section~\ref{sec:conclusions}.

\begin{table}
\setlength{\tabcolsep}{3.8pt}
\centering\footnotesize
\begin{tabular}{l cc cc cc}
\toprule
&
 \multirow{2}{*}{\begin{minipage}{0.9cm}\centering \#~Anno\-tators\end{minipage}} &
\multirow{2}{*}{\begin{minipage}{0.9cm}\centering Hours paid\end{minipage}} &
\multicolumn{2}{c}{Collected counts} &
\multicolumn{2}{c}{After filtering}
 \\ \cmidrule(lr){4-5} \cmidrule(lr){6-7}

& & & Textual & Visual & Textual & Visual \\
 \midrule

CZ & 9 &  \hphantom{1}89 & 1,242 & 596 & 1,080 & 456 \\
SK & 8 &  120 & 1,203 & 329 & \hphantom{1,}972 & 238 \\
UA & 9 & 110 & \hphantom{1,}889 & 740 & \hphantom{1,}755 & 403 \\
\bottomrule
\end{tabular}

\caption{Basic descriptive stats of the data collection process.}\label{tab:dataCollection}

\end{table}

\section{Data Collection}\label{sec:collection}

\begin{table}[t]
\centering\footnotesize
\setlength{\tabcolsep}{5pt}

\begin{tabular}{l r r r r c c}
\toprule
 & \multicolumn{4}{c}{Number of seed Wikipedia pages} & \multicolumn{2}{c}{\% of seed} \\
\cmidrule(lr){2-5} \cmidrule(lr){6-7}
 & \begin{minipage}{1.1cm}\centering Single-\\ language\end{minipage} & \begin{minipage}{1.0cm}\centering Match\\ location\end{minipage} & Total & \begin{minipage}{0.7cm}\centering \% of\\ Wiki \end{minipage} & Text & Visual \\
\midrule
CZ & 111k & 12k & 123k & 21.4 & 31.9 & 20.6 \\
SK & 36k & 4k & 40k & 15.7 & 32.6 & 15.3 \\
UA & 187k & 10k & 197k & 14.2 & 25.2 & 18.1 \\
\bottomrule
\end{tabular}
\caption{Number of seed Wikipedia pages per language in the pool. Single-language are pages that exist only in the local language, Matching location are pages about entities located in the respective country, and \% of Wikipedia shows what percentage of the local Wikipedia the pool represents. ``\% of seed'' denotes the percentage of dataset items created from seed articles, grouped by modality.}
\label{tab:wiki_pool_merged}
\end{table}


In this section, we describe our data collection methodology that relies on native speakers to identify regional-specific facts that are well-known within specific countries (Czechia, Slovakia, and Ukraine) but largely unknown outside these regions. 
The basic statistics of the data we collected are shown in Table~\ref{tab:dataCollection}.

The annotators were asked to find facts specific to their country (Czechia, Slovakia, Ukraine), i.e., facts that are commonly known locally but largely unknown outside the country.
To avoid overly obscure facts while maintaining regional specificity, we provided annotators with a guideline to consider whether the fact would be known by a substantial portion of the population (we suggested thinking of at least 10,000 people as a rough threshold). Annotators ultimately relied on their judgment of what constitutes regionally common knowledge.

We developed a web-based interface for collecting the data. In the interface, the annotators were presented with a Wikipedia page randomly selected from a pre-defined pool. The tool allows annotators to browse Wikipedia content, including clicking links navigating different Wikipedia pages or skipping the current page, redirecting the annotator to a random page from the seed pool.
The article pool consisted of Wikipedia pages about (1) entities that are located in the respective country in DBPedia \citep{mendes-etal-2012-dbpedia}, and (2) pages that only existed in the given language. Table~\ref{tab:wiki_pool_merged} shows the number of Wikipedia pages per language that were available in the pool.
Each annotator could only create a single annotation for each Wikipedia page.
The code for the annotation interface is available on GitHub.
\footnote{\url{https://github.com/ufal/regional-qa-annotation}}

If a suitable fact was found on a page, the annotators wrote a question and an answer in their language (Czech, Slovak, Ukrainian) and provided an English translation of both.
If the Wikipedia page had a photograph, the annotators were encouraged to also write down a question and an answer about the photograph (including the English translations).

All annotators are native speakers of the respective languages with sufficient knowledge of English (have language certificate and/or passed a compulsory university English exam), who were raised in the countries where these languages are spoken.  However, all annotators are currently university students living in the Czech Republic.
The annotators were compensated at a rate of 300 CZK ($\approx$ 12€) per hour, which is 2.4$\times$ the minimum wage and 1.2$\times$ the median wage in the Czech Republic as of 2024.

We manually verified \emph{all} questions and answers to ensure they adhered to the annotation guidelines. Questions were rejected if they: (1) concerned globally known facts, (2) were too obscure, (3) contained factual errors, or (4) had unclear phrasing. During this process, we corrected 8\% of the items (mostly typos and grammatical issues, in both languages) and discarded 22\% of the questions. All filtering decisions were made by research team members with domain expertise. Our manual inspection did not find any major issues in the quality of the English translations provided by the annotators.

\section{Dataset Description}\label{sec:dataset}

\begin{table}[t]
\centering\footnotesize
\setlength{\tabcolsep}{5.1pt}
\begin{tabular}{cl cc cc cc c }
\toprule

& & \multicolumn{2}{c}{\multirow{2}{*}{\begin{minipage}{2cm}\centering\vspace{2pt} Number of \\ instances\end{minipage}}} & \multicolumn{4}{c}{Avg. \# chars} & \multirow{3}{*}{\begin{minipage}{0.8cm}\centering Ans. \\ is \\ a full \\ sent.?\end{minipage}} \\ \cmidrule(lr){5-8}

& & & & \multicolumn{2}{c}{Local} & \multicolumn{2}{c}{English} &  \\ \cmidrule(lr){3-4} \cmidrule(lr){5-6} \cmidrule(lr){7-8}
& & Dev & Test & Q & A & Q & A \\ \midrule

\multirow{3}{*}{\rotatebox[origin=c]{90}{Textual}}
& CZ & 530 & 550 & 65 & 19 & 72 & 22 & \hphantom{1}4\% \\
& SK & 493 & 479 & 57 & 19 & 62 & 19 & \hphantom{1}7\% \\
& UA & 385 & 370 & 65 & 32 & 69 & 34 & 14\% \\ \midrule
\multirow{3}{*}{\rotatebox[origin=c]{90}{Visual}}
& CZ & 226 & 230  & 31 & 18 & 38 & 21 & \hphantom{1}2\% \\
& SK & 118 & 120  & 38 & 20 & 43 & 22 & \hphantom{1}9\%\\
& UA & 204 & 199 & 40 & 34 & 45 & 36 & 18\% \\ \bottomrule

\end{tabular}

\caption{Basic descriptive statistics of the dataset: dev and test split size, the average length of questions and answers in the number of characters, and proportion of answers that are a full sentence.}\label{tab:basic_stats}

\end{table}

\begin{table}[t]
\setlength{\tabcolsep}{4.5pt}
\centering\footnotesize
\begin{tabular}{ll rrr @{\hskip 12pt} rrr}
\toprule
& & \multicolumn{3}{c}{Textual} & \multicolumn{3}{c}{Visual} \\ \cmidrule(lr{12pt}){3-5} \cmidrule(l{0pt}r){6-8}
& & CZ & SK & UK & CZ & SK & UK \\ \midrule

\multirow{6}{*}{\rotatebox[origin=c]{90}{Categories}}
& Geography  & \Cat{39} & \Cat{42} & \Cat{32} & \Cat{60} & \Cat{50} & \Cat{39} \\
& Culture    & \Cat{27} & \Cat{23} & \Cat{19} & \Cat{21} & \Cat{24} & \Cat{20} \\
& Politics   & \Cat{5} & \Cat{9} & \Cat{12} & \Cat{3} & \Cat{9} & \Cat{9} \\
& History    & \Cat{21} & \Cat{12} & \Cat{26} & \Cat{10} & \Cat{6} & \Cat{20} \\
& Sports     & \Cat{5} & \Cat{8} & \Cat{5} & \Cat{2} & \Cat{3} & \Cat{5} \\
& Other      & \Cat{3} & \Cat{6} & \Cat{4} & \Cat{4} & \Cat{8} & \Cat{5} \\

\midrule

\multirow{4}{*}{\rotatebox[origin=c]{90}{Basic NER}}
& Location & \Ber{94} & \Ber{87} & \Ber{85} & \Ber{76} & \Ber{60} & \Ber{57} \\
& Person & \Ber{46} & \Ber{44} & \Ber{50} & \Ber{32} & \Ber{39} & \Ber{43} \\
& Organization & \Ber{33} & \Ber{41} & \Ber{45} & \Ber{11} & \Ber{19} & \Ber{31} \\
& Date/Year & \Ber{8} & \Ber{7} & \Ber{11} & \Ber{1} & \Ber{0} & \Ber{5} \\

\midrule

\multirow{20}{*}{\rotatebox[origin=c]{90}{Fine-grained named entities}}

& River & \Ner{9} & \Ner{8} & \Ner{7} & \Ner{4} & \Ner{3} & \Ner{1} \\
& Mountain & \Ner{14} & \Ner{17} & \Ner{6} & \Ner{10} & \Ner{8} & \Ner{6} \\
& City/Town & \Ner{56} & \Ner{48} & \Ner{65} & \Ner{43} & \Ner{40} & \Ner{42} \\
& Village & \Ner{11} & \Ner{13} & \Ner{5} & \Ner{5} & \Ner{9} & \Ner{2} \\
& Movie & \Ner{3} & \Ner{1} & \Ner{0} & \Ner{0} & \Ner{0} & \Ner{0} \\
& TV show & \Ner{5} & \Ner{7} & \Ner{3} & \Ner{0} & \Ner{0} & \Ner{1} \\
& Book & \Ner{7} & \Ner{3} & \Ner{3} & \Ner{2} & \Ner{0} & \Ner{1} \\
& Song & \Ner{1} & \Ner{2} & \Ner{3} & \Ner{0} & \Ner{0} & \Ner{1} \\
& Actor & \Ner{6} & \Ner{3} & \Ner{3} & \Ner{1} & \Ner{1} & \Ner{1} \\
& Writer & \Ner{9} & \Ner{6} & \Ner{8} & \Ner{3} & \Ner{3} & \Ner{5} \\
& Musician & \Ner{3} & \Ner{6} & \Ner{4} & \Ner{2} & \Ner{10} & \Ner{1} \\
& Other artist & \Ner{5} & \Ner{6} & \Ner{8} & \Ner{2} & \Ner{5} & \Ner{4} \\
& Politian & \Ner{13} & \Ner{13} & \Ner{24} & \Ner{3} & \Ner{11} & \Ner{14} \\
& Sportsperson & \Ner{2} & \Ner{5} & \Ner{2} & \Ner{1} & \Ner{5} & \Ner{2} \\
& Other person & \Ner{32} & \Ner{24} & \Ner{32} & \Ner{29} & \Ner{27} & \Ner{33} \\
& Sports event & \Ner{5} & \Ner{9} & \Ner{5} & \Ner{0} & \Ner{1} & \Ner{3} \\
& Political party & \Ner{3} & \Ner{5} & \Ner{5} & \Ner{0} & \Ner{0} & \Ner{1} \\
& Other org. & \Ner{21} & \Ner{27} & \Ner{36} & \Ner{6} & \Ner{12} & \Ner{22} \\
& Product/Brand & \Ner{11} & \Ner{8} & \Ner{13} & \Ner{8} & \Ner{2} & \Ner{14} \\
& Company & \Ner{4} & \Ner{2} & \Ner{5} & \Ner{0} & \Ner{1} & \Ner{3} \\

\bottomrule

\end{tabular}

\caption{Percentage of questions belonging to six mutually exclusive main categories (upper part) and a proportion of question-answer pairs containing named entities of the listed types with different granularity (middle and lower parts).}\label{tab:semantic_stats}

\end{table}

\begin{table}

\footnotesize\centering
\setlength{\tabcolsep}{3.5pt}
\begin{tabular}{l cccccccc @{\hskip 10pt} c @{\hskip 10pt}ccc}
\toprule

    \multirow{2}{*}{\rotatebox{90}{Location\quad\quad}} & \multicolumn{8}{c}{content} &  & \multicolumn{3}{c}{medium} \\
    \cmidrule(lr){2-9} \cmidrule(lr){11-13}
& \rotatebox{90}{portrait} & \rotatebox{90}{building} & \rotatebox{90}{city} & \rotatebox{90}{nature} & \rotatebox{90}{statue} & \rotatebox{90}{food} & \rotatebox{90}{technology} & \rotatebox{90}{other} & \rotatebox{90}{has text} & \rotatebox{90}{color photo} & \rotatebox{90}{BW photo} & \rotatebox{90}{painting} \\
\midrule
CZ  & \Ner{23} & \Ner{57} & \Ner{30} & \Ner{8} & \Ner{8} & \Ner{2} & \Ner{1} & \Ner{4} & \Ner{18} & \Ner{86} & \Ner{10} & \Ner{4} \\
SK  & \Ner{31} & \Ner{52} & \Ner{15} & \Ner{5} & \Ner{4} & \Ner{1} & \Ner{0} & \Ner{6} & \Ner{25} & \Ner{92} & \Ner{7} & \Ner{2} \\
UA  & \Ner{25} & \Ner{36} & \Ner{27} & \Ner{5} & \Ner{4} & \Ner{3} & \Ner{1} & \Ner{21} & \Ner{23} & \Ner{79} & \Ner{10} & \Ner{10} \\
\bottomrule
\end{tabular}

\caption{Percentage of images in the dataset belonging to the listed categories based on the visual content.}\label{tab:image_categories}

\end{table}

\begin{table*}[t!]
\centering\footnotesize
\begin{tabular}{l p{6cm} p{6cm}}
\toprule
Image & Question & Answer \\ \midrule

\multirow{2}{*}{\includegraphics[trim={0cm 13cm 0cm 0cm}, clip, width=2.5cm]{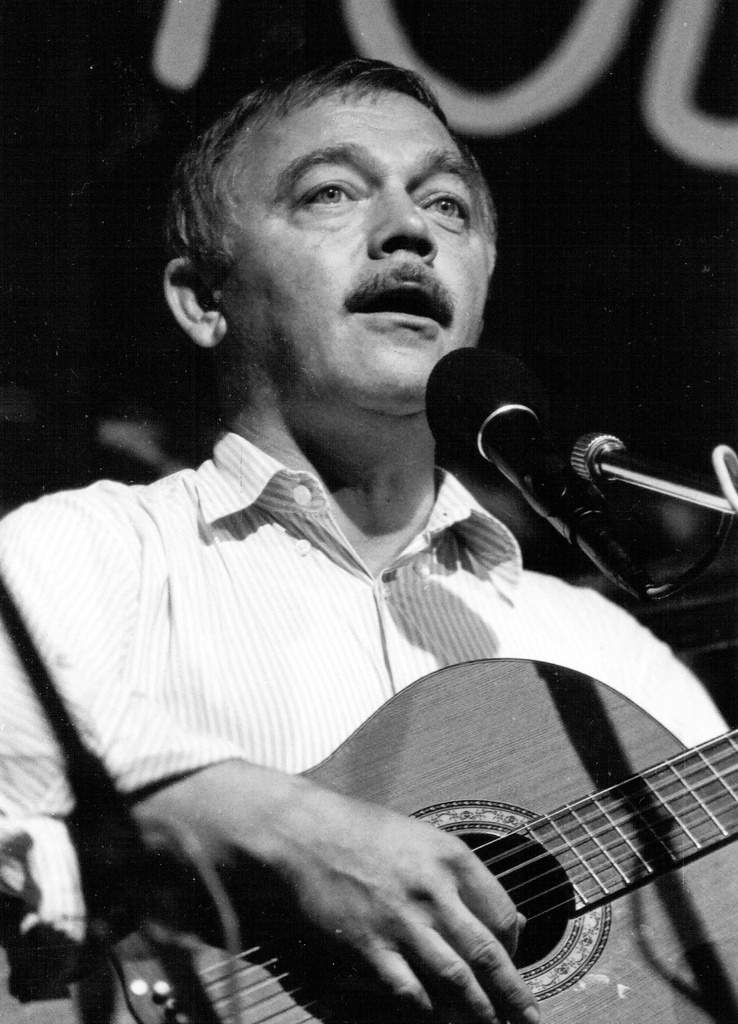}} & Jak se jmenuje zpěvák na obrázku? & Karel Kryl \\
& \it What is the name of the singer in the picture? & \it Karel Kryl \\[45pt] \midrule

\multirow{2}{*}{\includegraphics[width=2.5cm]{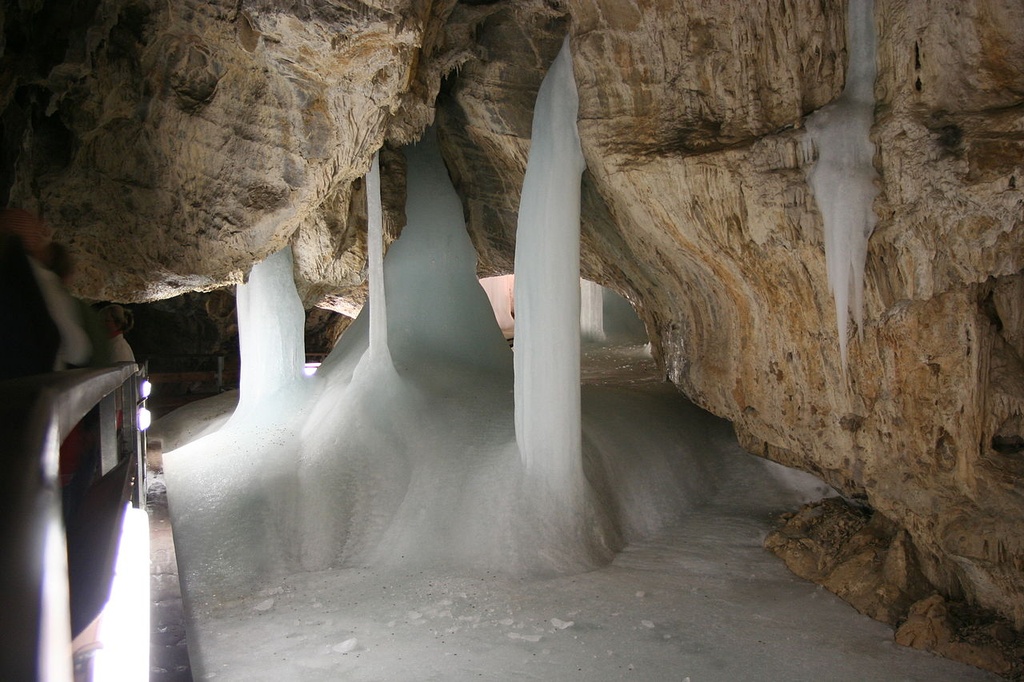}} & Ktorá slovenská jaskyňa je na obrázku? & Demänovská ľadová jaskyňa \\
& \it Which Slovak cave is in the picture? & \it Demänovská Ice Cave \\ [28pt] \midrule

\multirow{2}{*}{\includegraphics[trim={0cm 6cm 0cm 1cm}, clip, width=2.5cm]{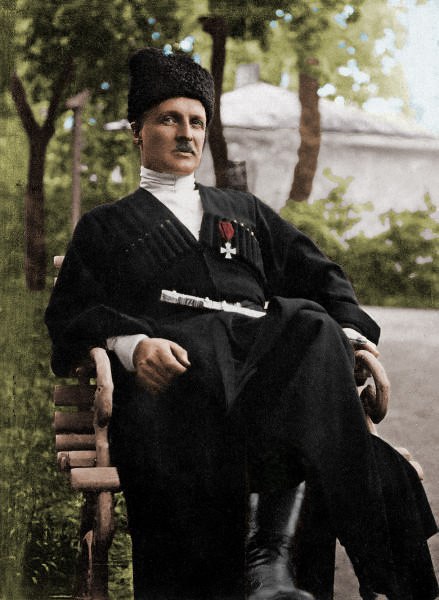}} & \cyr{Хто зображений на фотографії?} & \cyr{На фотографії зображена постать Гетьмана Павла Скоропадського.} \\
 & \it Who is shown in the photo? & \it The photo shows the figure of Hetman Pavel Skoropadskyi. \\ [33pt]

\bottomrule
\end{tabular}

\caption{Examples of collected Czech, Slovak and Ukrainian questions and answers. More examples from both modalities in Tables~\ref{tab:textual_extra} and~\ref{tab:example_extra} in the Appendix.}\label{tab:example}
\end{table*}

Table~\ref{tab:basic_stats} lists characteristics of the dataset.
For each country and question type, the dataset is divided into a development and test split.
Each example in the dataset consists of the following, manually annotated features:
\begin{itemize}

  \item the Wikipedia page URL,

  \item a question in the local language,

  \item a answer in the local language,

  \item the same question translated into English,

  \item the corresponding answer in English, and

  \item the image data (visual modality only).

\end{itemize}
We show examples of the collected data in the visual modality part in Table~\ref{tab:example}, with additional examples from both modalities in Tables~\ref{tab:textual_extra} and~\ref{tab:example_extra} in the appendix.
We release the dataset on Hugging Face Hub.%
\footnote{\url{https://huggingface.co/datasets/ufal/cus-qa}}

The average length of questions in the textual modality part is around 60 characters in all languages; the English translations are slightly longer. The questions in the visual modality part are shorter because some context is already provided by the image. In both modalities, the answers are, on average, 20 characters long in Czech and Slovak but over 30 in Ukrainian. Most of the answers are short noun phrases. In Czech, only a small portion of the answers are full sentences; it is slightly more in Slovak, and almost one-sixth of the answers in Ukrainian.

Besides the manually created features, we augment each example with automatic translations into the other two local languages, using Claude 3.5 Sonnet (best scoring system in WMT 2024; \citealp{kocmi-etal-2024-findings}), which we later use for cross-lingual comparison of the models (\S~\ref{ssec:xling_diff}).
To verify the translation quality, we randomly sampled 100 question-answer pairs in three translation directions and compared Google Translation, GPT-4o, and Claude 3.5 Sonet (Table~\ref{tab:manual_mt}). Because the questions and answers are quite simple texts, the translation was perfect in the vast majority of cases, with only minor issues in the rest.

\begin{table}

\centering\footnotesize
\setlength{\tabcolsep}{2.8pt}
\begin{tabular}{l cccccc}
\toprule
Translator       & sk-cs & uk-cs & cs-sk & uk-sk & cs-uk & sk-uk \\ \midrule
Google Translate &   78  &   65  &   57  &   75  &   73  &   76  \\
GPT-4o           &   98  &   86  &   94  &   82  &   88  &   87  \\
Claude 3.5 Sonet &   98  &   88  &   97  &   84  &   90  &   90  \\
\bottomrule
\end{tabular}

\caption{Comparison of Google Translation, GPT-4o and Claude 3.5 Sonet. Sentences were evaluated by a single annotator, indicating whether the translation can be considered perfect. Low performance of Google Translation is due to pivoting via English, which leads to incorrect transliteration and English names in the translations.}\label{tab:manual_mt}

\end{table}

For a quantitative summary of the dataset content, we use LLaMA 3.3 70B \citep{dubey2024llama} to assign one of the following categories to each example: Geography, Culture, Politics, History, Sports, and Other. 
We treat these categories as mutually exclusive, even though sometimes, the classification may be ambiguous (e.g., History vs. Politics). The category statistics are shown in Table~\ref{tab:semantic_stats}. In both textual and visual questions, geography was the most frequent category, followed by culture and history.

Additionally, we use GliNER \citep{zaratiana-etal-2024-gliner} on the English translation of the question-answer pairs and we include the extracted statistics in Table~\ref{tab:semantic_stats}. GliNER allows defining custom entities based on prompts, so besides the basic entities (location, person, organization, date), we defined our own set of fine-grained entities that better describe the content of the dataset.
The overview of the named entities shows that the vast majority of question-answer pairs contain a location, with city names being its most frequent subtype. Roughly 40\% of the questions contain the name of a person. According to GliNER, the most frequent subcategory of names belongs to politicians. However, this might be inaccurate, as we apply an English-language recognizer to non-English names.

Similarly for the visual part of the dataset, we used LLaMA 3.2 11B Vision to categorize the images based on whether they primarily capture faces of people (portraits), a building, a town or city, natural scenery, a statue, food, or technological artifacts (such as cars and tools). Photos that do not contain any of these are categorized as other. Further, we detect if there is text in the photo. Finally, we classify whether it is a color photo, black-and-white photo or a different medium (in most cases a drawing or a painting). The statistics are in Table~\ref{tab:image_categories}.


CUS-QA is designed for evaluation purposes only. The dataset size (1,536 Czech, 1,210 Slovak, and 1,158 Ukrainian question-answer pairs) is intentionally limited to serve as a test set for assessing model capabilities on regional knowledge rather than as training data. 

\section{Baselines \& Evaluation}\label{sec:evaluation}

We evaluate several state-of-the-art open-weight models on CUS-QA to establish baseline performance and examine cross-lingual consistency. Our experiments test how well current LLMs perform on regional questions across both textual and visual modalities, and whether models maintain consistent performance when questions are asked in different languages.

We use human evaluation to analyze the reliability of automatic metrics for open-ended QA. We compare various evaluation approaches with human judgments to determine which metrics best capture answer quality across different languages and modalities.

\subsection{Baselines}

As baselines, we test zero-shot generation performance of several widely used open-weight LLMs in each modality setting.
In the text-only setup, we evaluate the following models: LLaMA 3.1 8B, LLaMA 3.3 70B \citep{dubey2024llama}, Mistral v0.3 7B \citep{jiang2023mistral} and EuroLLM 7B \citep{martins2024eurollm}. For visual QA, we test mBLIP \citep{geigle2023mblip}, LLaMA 3.2 and 4 Vision, Maya \citep{alam2024mayainstructionfinetunedmultilingual}, Gemma 3 \citep{gemmateam2025gemma3technicalreport} and idefics \citep{laurençon2023obelicsopenwebscalefiltered}. 

Based on preliminary experiments, we decided to use an English system prompt and instruct the model to answer in the language of the questions. See Appendix~\ref{app:prompts} for the exact prompt formulation.
For decoding, we used nucleus sampling with a nucleus of 0.9 in all setups, and we only sampled one answer per question and model.

\subsection{Human Evaluation}

We manually evaluate the model outputs using the following binary criteria: correctness, truthfulness, relevance, and coherence.
Our aim was to design a small set of error categories that best describe what we observed in the generated answers.

\paragraph{Correctness.}

In the main evaluation criterion, the annotators decided if the generated answer is correct, given the reference answer, using their subjective judgment. If the answer contained minor hallucinations or misleading information that do not substantially change the overall meaning of the answer, it is still considered correct.

\paragraph{Truthfulness.}

The annotators verified whether the entire answer is objectively true. This is separate from whether the answer addresses the question. An answer can be factually true but incorrect if it does not answer the question. We also allow saying that the answer is correct if it contains irrelevant, untrue statements. If the model refuses to answer, it is annotated as not true.

\paragraph{Relevance.}

Here, the annotators judged whether the answer contains only the information the question asks for and nothing else, is appropriately specific, and is not too general.

\paragraph{Coherence.}

A coherent answer is grammatically correct and in the correct language. This is judged regardless of the factuality or relevance.

\vspace{0.8\baselineskip}
\noindent
These categories allow us to define different levels of strictness in how we judge the answers: In the permissive case, we only consider the correctness criterion, and in the strict case, we want all criteria to be fulfilled. Each answer was independently evaluated by two annotators. We consider a criterion to be fulfilled when both annotators agree.

The annotators in the evaluation campaign received the same compensation as in the data collection phase (\S~\ref{sec:collection}). Some of the annotators for data collection were also involved in human evaluation. All of the annotators for this phase are native speakers of the relevant language. The total time spent on manual evaluation was 106 hours.

Inter-annotator agreement (Table~\ref{tab:agreement}), measured by Cohen's $\kappa$, shows generally strong consistency between annotators across most evaluation criteria. The agreement is highest for correctness and truthfulness (average $\kappa = 0.84$ and $0.86$ respectively). It is lower for relevance ($\kappa = 0.70$) and coherence ($\kappa = 0.80$), likely reflecting the more subjective nature of these judgments.

\subsection{Automatic Metrics}

We evaluate the generated answers using the following metrics:
BLEU \citep{papineni-etal-2002-bleu}, chrF \citep{popovic-2015-chrf}, ROUGE-L \citep{lin-2004-rouge}, BERTScore \citep{zhang2020bertscore} with XLM-R \citep{conneau-etal-2020-unsupervised} as an underlying model, BLEURT \citep{sellam-etal-2020-bleurt}.

Additionally, we use LLM as a judge \citep{zheng2023judge}. We use LLaMA 3.3 70B, a general-purpose model, and M-Prometheus that was finetuned to be used as a judge in multilingual scenarios.
The judge prompt contains the question, the reference answer, the model output, and the question of whether the answer is correct or not (see Appendix~\ref{app:prompts} for the full prompt). The LLM-based judges produce a binary value (correct/incorrect) and we report the ratio of correct answers in the dataset.

\subsection{Retrieval Augmented Generation}

Because the dataset design ensures that all answers are sourced from the local Wikipedia, we conduct additional experiments on the textual part of the dataset using retrieval-augmented generation (RAG).

We used the FineWiki dataset \citep{penedo2025finewiki} to build the RAG index for each local Wikipedia. We split each document into overlapping chunks, use the Multilingual-E5-Large text embeddings 
\citep{wang2024efive} to get the vector representations of the chunks and store the vectors in the index. In all experiments, we set the chunk size to 512 tokens, with 10\% overlap between neighboring chunks of the same document. We use the Faiss library for building the index and retrieval \citep{douze2024faiss}.

During decoding, we embed the question using the same model, and retrieve the top 3 most similar vectors and the respective chunks from the index, and include the retrieved data in the prompt (see Appendix~\ref{app:prompts} for the exact prompt formulation).

\begin{table}[t]


\footnotesize\centering
\begin{tabular}{lll cccc}
\toprule
& & &
\begin{minipage}{9.4mm}\centering Correct \end{minipage}  &
\begin{minipage}{9.4mm}\centering True    \end{minipage}  &
\begin{minipage}{9.4mm}\centering \textls[-64]{Relevant}\end{minipage}  &
\begin{minipage}{9.4mm}\centering \textls[-85]{Coherent}\end{minipage}  \\
\midrule

\multirow{6}{*}{\rotatebox{90}{Text}}
& \multirow{2}{*}{CZ}
  & cs & \agreement{.91} & \agreement{.90} & \agreement{.84} & \agreement{.76} \\
& & en & \agreement{.90} & \agreement{.86} & \agreement{.72} & \agreement{.84} \\
\cmidrule{2-7}
& \multirow{2}{*}{SK}
  & sk & \agreement{.90} & \agreement{.88} & \agreement{.72} & \agreement{.69} \\
& & en & \agreement{.90} & \agreement{.85} & \agreement{.61} & \agreement{.85} \\
\cmidrule{2-7}
& \multirow{2}{*}{UA}
  & uk & \agreement{.84} & \agreement{.83} & \agreement{.79} & \agreement{.88} \\
& & en & \agreement{.78} & \agreement{.79} & \agreement{.58} & \agreement{.91} \\
\midrule
\multirow{6}{*}{\rotatebox{90}{Visual}}
& \multirow{2}{*}{CZ}
  & cs & \agreement{.65} & \agreement{.83} & \agreement{.55} & \agreement{.64} \\
& & en & \agreement{.58} & \agreement{.85} & \agreement{.47} & \agreement{.81} \\
\cmidrule{2-7}
& \multirow{2}{*}{SK}
  & sk & \agreement{.90} & \agreement{.91} & \agreement{.83} & \agreement{.66} \\
& & en & \agreement{.91} & \agreement{.91} & \agreement{.88} & \agreement{.83} \\
\cmidrule{2-7}
& \multirow{2}{*}{UA}
  & uk & \agreement{.91} & \agreement{.83} & \agreement{.75} & \agreement{.83} \\
& & en & \agreement{.90} & \agreement{.82} & \agreement{.65} & \agreement{.93} \\
\midrule
\multicolumn{3}{l}{Average $\kappa$} &
\agreement{.84} & \agreement{.86} & \agreement{.70} & \agreement{.80} \\
\bottomrule
\end{tabular}

\caption{Inter-annotator agreement for all evaluation criteria for two annotators measured by Cohen's $\kappa$.}\label{tab:agreement}

\end{table}

\begin{table}[t]
\footnotesize\centering
\setlength{\tabcolsep}{4.7pt}
\begin{tabular}{lcc cc cc cc}
\toprule
& & & \multicolumn{2}{c}{Human dev} & \multicolumn{2}{c}{Auto. dev} & \multicolumn{2}{c}{Auto. test} \\ \cmidrule(lr){4-5} \cmidrule(lr){6-7} \cmidrule(lr){8-9}

& &
 & \begin{minipage}{6.0mm}\centering\textls[-0]{OK}\end{minipage}
 & \begin{minipage}{6.0mm}\centering Perf.\end{minipage}

 & \begin{minipage}{6.0mm}\centering chrF\end{minipage}
 & \begin{minipage}{6.0mm}\centering LLM\end{minipage}

 & \begin{minipage}{6.0mm}\centering chrF\end{minipage}
 & \begin{minipage}{6.0mm}\centering LLM\end{minipage}
\\
\midrule


\multirow{6}{*}{\rotatebox{90}{Textual}} & \multirow{2}{*}{CZ} & cs & \human{58.7} & \human{53.4} & \chrf{42.4} & \llamascore{57.4} & \chrf{43.7} & \llamascore{60.0} \\
 &  & en & \human{52.5} & \human{44.7} & \chrf{37.8} & \llamascore{56.0} & \chrf{37.5} & \llamascore{63.1} \\
\cmidrule{2-9}
 & \multirow{2}{*}{SK} & sk & \human{46.2} & \human{40.8} & \chrf{39.2} & \llamascore{45.4} & \chrf{35.2} & \llamascore{47.7} \\
 &  & en & \human{45.8} & \human{40.4} & \chrf{32.5} & \llamascore{47.5} & \chrf{32.6} & \llamascore{48.3} \\
\cmidrule{2-9}
 & \multirow{2}{*}{UA} & uk & \human{51.9} & \human{46.2} & \chrf{42.4} & \llamascore{57.4} & \chrf{42.6} & \llamascore{52.2} \\
 &  & en & \human{47.8} & \human{43.1} & \chrf{37.8} & \llamascore{56.0} & \chrf{37.0} & \llamascore{54.9} \\
\midrule
\multirow{6}{*}{\rotatebox{90}{Visual}} & \multirow{2}{*}{CZ} & cs & \human{14.6} & \human{11.1} & \chrf{21.3} & \llamascore{14.6} & \chrf{21.3} & \llamascore{14.6} \\
 &  & en & \human{13.3} & \human{8.4} & \chrf{15.5} & \llamascore{15.9} & \chrf{15.5} & \llamascore{15.9} \\
\cmidrule{2-9}
 & \multirow{2}{*}{SK} & sk & \human{23.7} & \human{20.3} & \chrf{24.4} & \llamascore{18.6} & \chrf{24.4} & \llamascore{18.6} \\
 &  & en & \human{19.5} & \human{12.7} & \chrf{16.6} & \llamascore{16.9} & \chrf{20.4} & \llamascore{17.5} \\
\cmidrule{2-9}
 & \multirow{2}{*}{UA} & uk & \human{29.9} & \human{29.9} & \chrf{21.1} & \llamascore{26.0} & \chrf{20.7} & \llamascore{15.1} \\
 &  & en & \human{17.2} & \human{17.2} & \chrf{19.1} & \llamascore{20.1} & \chrf{21.7} & \llamascore{23.6} \\

 \bottomrule

\end{tabular}

\caption{Results of the strongest baselines for textual (LLaMA 3.3 70B Instruct) and visual (LLaMA 4 Scout 17B 16E Instruct) QA tasks. Human evaluation includes factual correctness (OK) and full correctness that also includes truthfulness, relevance, and coherence (Perfect). Additionally, we report the chrF score and LLaMA 3.3 70B as a judge for development and test data. Complete results of all tested models are in the Appendix in Tables~\ref{tab:detailed_baselines_textual} and~\ref{tab:detailed_baselines_visual}.}\label{tab:baselines}

\end{table}

\begin{table}[]
    \centering\footnotesize
    \setlength{\tabcolsep}{3.5pt}
    \begin{tabular}{l cc cc cc}
    \toprule
    \multirow{2}{*}{Model} & \multicolumn{2}{c}{CZ} & \multicolumn{2}{c}{SK} & \multicolumn{2}{c}{UA} \\ \cmidrule(lr){2-3} \cmidrule(lr){4-5} \cmidrule(lr){6-7}
    & cs & en & sk & en & uk & en \\ \midrule

LLaMA 3.3 70B Ins. & \llamaallscore{60.0} & \llamaallscore{63.1} & \llamaallscore{47.7} & \llamaallscore{48.3} & \llamaallscore{52.2} & \llamaallscore{54.9} \\
LLaMA 3.1 8B Ins. & \llamaallscore{29.6} & \llamaallscore{36.5} & \llamaallscore{19.6} & \llamaallscore{25.6} & \llamaallscore{25.7} & \llamaallscore{34.1} \\
EuroLLM 9B Ins. & \llamaallscore{39.3} & \llamaallscore{43.1} & \llamaallscore{30.8} & \llamaallscore{30.2} & \llamaallscore{26.8} & \llamaallscore{28.4} \\
Mistral 7B Ins. v0.3 & \llamaallscore{22.9} & \llamaallscore{30.4} & \llamaallscore{14.4} & \llamaallscore{22.3} & \llamaallscore{24.6} & \llamaallscore{27.8} \\
LLaMA 4 Scout Ins. & \llamaallscore{41.8} & \llamaallscore{45.6} & \llamaallscore{34.6} & \llamaallscore{34.4} & \llamaallscore{45.7} & \llamaallscore{45.4} \\

\it Gemini 2.5 Flash & \llamaallscore{78.0} & \llamaallscore{75.8} & \llamaallscore{64.2} & \llamaallscore{61.3} & \llamaallscore{68.9} & \llamaallscore{63.0} \\
\it Gemini 2.5 Fl.-Lite & \llamaallscore{60.0} & \llamaallscore{54.2} & \llamaallscore{47.9} & \llamaallscore{43.3} & \llamaallscore{54.3} & \llamaallscore{51.9} \\
\it GPT-5\tiny-2025-08-07 & \llamaallscore{83.1} & \llamaallscore{80.7} & \llamaallscore{73.3} & \llamaallscore{70.2} & \llamaallscore{73.2} & \llamaallscore{72.2} \\
\it GPT-5-mini\tiny-2025-08-07 & \llamaallscore{67.5} & \llamaallscore{70.7} & \llamaallscore{52.3} & \llamaallscore{50.4} & \llamaallscore{61.4} & \llamaallscore{59.7} \\

\midrule

LLaMA 3.2 11B Ins. & \llamascore{12.6} & \llamascore{14.3} & \llamascore{11.7} & \llamascore{15.0} & \llamascore{11.6} & \llamascore{12.1} \\
LLaMA 4 Scout Ins. & \llamascore{18.7} & \llamascore{18.3} & \llamascore{15.0} & \llamascore{17.5} & \llamascore{15.1} & \llamascore{23.6} \\
maya & \llamascore{1.7} & \llamascore{3.0} & \llamascore{1.7} & \llamascore{2.5} & \llamascore{2.5} & \llamascore{4.0} \\
idefics & \llamascore{4.8} & \llamascore{7.4} & \llamascore{6.7} & \llamascore{9.2} & \llamascore{6.0} & \llamascore{8.5} \\
gemma3 & \llamascore{15.2} & \llamascore{13.9} & \llamascore{11.7} & \llamascore{13.3} & \llamascore{16.6} & \llamascore{12.1} \\

\it Gemini 2.5 Flash & \llamascore{39.6} & \llamascore{37.0} & \llamascore{38.3} & \llamascore{34.2} & \llamascore{41.2} & \llamascore{40.7} \\
\it Gemini 2.5 Fl.-Lite & \llamascore{30.9} & \llamascore{27.4} & \llamascore{26.7} & \llamascore{24.2} & \llamascore{27.1} & \llamascore{27.6} \\
\it GPT-5\tiny-2025-08-07 & \llamascore{46.5} & \llamascore{42.6} & \llamascore{44.2} & \llamascore{41.7} & \llamascore{37.2} & \llamascore{36.2} \\
\it GPT-5-mini\tiny-2025-08-07 & \llamascore{33.0} & \llamascore{38.7} & \llamascore{24.2} & \llamascore{26.7} & \llamascore{26.1} & \llamascore{31.7} \\

    \bottomrule
    \end{tabular}

    \caption{Test data results for all models (including commercial ones) evaluated by LLaMA 3.3 70B as a judge. Textual question are in the upper part of the table, visual questions are in the lower part. Commercial models are in italics.}
    \label{tab:commercial}
\end{table}

\begin{table*}

\footnotesize\centering
\setlength{\tabcolsep}{4pt}
\newcolumntype{C}{>{\centering\arraybackslash}p{7mm}}
\begin{tabular}{l *{14}{C}}
\toprule
 & \multicolumn{7}{c}{Textual} & \multicolumn{7}{c}{Visual} \\
\cmidrule(lr){2-8} \cmidrule(lr){9-15}
 & BLEU & chrF & RG & BScr & BLRT & LLM & M-P & BLEU & chrF & RG & BScr & BLRT & LLM & M-P \\
 \midrule
Correct    & \Co{.733} & \Co{.867} & \Co{.833} & \Co{.817} & \Co{.883} & \Co{.950} & \Co{950} & \Co{.667} & \Co{.550} & \Co{.761} & \Co{-.267} & \Co{.817} & \Co{1.000} & \Co{.962} \\
+ true     & \Co{.753} & \Co{.879} & \Co{.845} & \Co{.829} & \Co{.879} & \Co{.970} & \Co{970} & \Co{.667} & \Co{.550} & \Co{.761} & \Co{-.267} & \Co{.817} & \Co{1.000} & \Co{.962} \\
+ relevant & \Co{.783} & \Co{.917} & \Co{.883} & \Co{.867} & \Co{.917} & \Co{.967} & \Co{967} & \Co{.667} & \Co{.550} & \Co{.761} & \Co{-.267} & \Co{.817} & \Co{1.000} & \Co{.962} \\
+ coherent & \Co{.783} & \Co{.917} & \Co{.883} & \Co{.867} & \Co{.917} & \Co{.967} & \Co{967} & \Co{.629} & \Co{.695} & \Co{.773} & \Co{-.215} & \Co{.829} & \Co{ .920} & \Co{.858} \\
\bottomrule
\end{tabular}

\caption{System-level Spearman correlations of the automatic evaluation metrics (BLEU, chrF, ROUGE-L, BERTScore, BLEURT, LLaMA 3.3 70B as a judge, M-Prometheus as a judge) with increasing levels of strictness of human evaluation. The presented numbers are averages of Spearman correlations computed separately for each country and language. The breakout for specific languages is in Table~\ref{tab:system_level_individual} in the Appendix. Pearson correlations are in Table~\ref{tab:appendix_pearson} in the Appendix.} %
\label{tab:system_level}

\end{table*}

\begin{table*}


\footnotesize\centering
\setlength{\tabcolsep}{4pt}
\newcolumntype{D}{>{\centering\arraybackslash}p{9mm}}
\begin{tabular}{l *{10}{D}}
\toprule
 & \multicolumn{5}{c}{Textual} & \multicolumn{5}{c}{Visual} \\
\cmidrule(lr){2-6} \cmidrule(lr){7-11}
 & chrF & RG & BLRT & LLM & M-P & chrF & RG & BLRT & LLM & M-P \\
 \midrule
Correct & \R{.594} & \R{.477} & \R{.552} & \R{.790} & \R{.707} & \R{.409} & \R{.370} & \R{.450} & \R{.807} & \R{.699} \\
+ true & \R{.585} & \R{.472} & \R{.543} & \R{.775} & \R{.703} & \R{.408} & \R{.370} & \R{.449} & \R{.804} & \R{.696} \\
+ relevant & \R{.567} & \R{.457} & \R{.524} & \R{.743} & \R{.683} & \R{.410} & \R{.370} & \R{.452} & \R{.797} & \R{.715} \\
+ coherent & \R{.548} & \R{.448} & \R{.511} & \R{.715} & \R{.665} & \R{.402} & \R{.376} & \R{.423} & \R{.710} & \R{.654} \\
\bottomrule
\end{tabular}

\caption{Answer-level point-wise biserial correlation of human judgment with automatic metrics, break-out per language is in Table~\ref{tab:answer_level_individual} in the Appendix.} %
\label{tab:answer_level}

\end{table*}

\begin{table*}
\centering\footnotesize

\begin{minipage}{0.06\textwidth}
\centering
Legend \\
\confmatlegend{}
\end{minipage}\quad
\begin{minipage}{0.06\textwidth}
\centering
Average \\
\confmat{29}{4}{5}{61}
\end{minipage}\quad
\begin{minipage}{0.01\textwidth}
\rotatebox{90}{\textbf{Textual}}
\end{minipage}~
\begin{minipage}{0.06\textwidth}
\centering
\vphantom{g}CZ cs \\
\confmat{34}{3}{2}{61}
\end{minipage}
\begin{minipage}{0.06\textwidth}
\centering
\vphantom{g}CZ en \\
\confmat{30}{5}{8}{57}
\end{minipage}
\begin{minipage}{0.06\textwidth}
\centering
\vphantom{g}SK sk \\
\confmat{26}{3}{2}{69}
\end{minipage}
\begin{minipage}{0.06\textwidth}
\centering
\vphantom{g}SK en \\
\confmat{25}{6}{7}{62}
\end{minipage}
\begin{minipage}{0.06\textwidth}
\centering
\vphantom{g}UA uk \\
\confmat{32}{3}{3}{63}
\end{minipage}
\begin{minipage}{0.06\textwidth}
\centering
\vphantom{g}UA en \\
\confmat{28}{7}{9}{56}
\end{minipage}\quad
\begin{minipage}{0.01\textwidth}
\rotatebox{90}{\textbf{Visual}}
\end{minipage}~
\begin{minipage}{0.06\textwidth}
\centering
\vphantom{g}CZ cs \\
\confmat{7}{1}{2}{90}
\end{minipage}
\begin{minipage}{0.06\textwidth}
\centering
\vphantom{g}CZ en \\
\confmat{7}{1}{3}{89}
\end{minipage}
\begin{minipage}{0.06\textwidth}
\centering
\vphantom{g}SK sk \\
\confmat{9}{4}{1}{86}
\end{minipage}
\begin{minipage}{0.06\textwidth}
\centering
\vphantom{g}SK en \\
\confmat{9}{2}{1}{88}
\end{minipage}
\begin{minipage}{0.06\textwidth}
\centering
\vphantom{g}UA uk \\
\confmat{13}{4}{2}{82}
\end{minipage}
\begin{minipage}{0.06\textwidth}
\centering
\vphantom{g}UA en \\
\confmat{14}{1}{4}{81}
\end{minipage}

\caption{Averaged confusion matrices comparing using LLaMA 3.3 70B Instruct as judge and human evaluation of correctness. The per-model breakout is in Table~\ref{tab:confusion_appendix} in the appendix.}\label{tab:confusion}

\end{table*}

\section{Results}
\label{sec:results}

We present the performance of the baseline models on open-ended regional question answering on CUS-QA (\S~\ref{ssec:baseline_res}), analyze the correlation between automatic metrics and human judgments (\S~\ref{ssec:hum_judgment}), and examine cross-lingual consistency in model capabilities (\S~\ref{ssec:xling_diff}). We include assessment of the robustness of the results to prompt variations (\S~\ref{ssec:promptvars}), and proof-of-concept experiments with retrieval-augmented generation (\S~\ref{ssec:ragresults}).

\subsection{Baseline Results}\label{ssec:baseline_res}

We report the strongest baseline results in Table~\ref{tab:baselines}, including results of manual evaluation and best-correlating automatic metrics (chrF, LLM as a judge) on the development and test data in both modalities.
Table~\ref{tab:commercial} shows LLM-as-a-judge evaluation for all tested models, including recent commercial systems.

\paragraph{Textual QA.}

The best-scoring model for textual QA was LLaMA 3.3, which outperformed the other models by a large margin, often 15--20 percentage points in accuracy compared to the others, and performed consistently well across languages and regions. There is a gap between focusing on correctness only, versus considering all evaluation criteria. This shows that models often add irrelevant or hallucinated information to their answers.
Bigger commercial models (GPT-5 and Gemini 2.5 Flash) outperform the best open-weight model by a large margin. Faster versions have a similar score to the best open-weight model.

Results of all textual models, including all metrics and human evaluation breakdown, are presented in Table~\ref{tab:detailed_baselines_textual} in the Appendix.
All models show differences in performance across languages. Mistral performs much worse in Slovak than in other languages. LLaMA 3.1 also struggles with Slovak, and like Mistral, it answers questions about Slovakia better in English than in Slovak.
EuroLLM, which focuses on European languages, outperforms both models of similar size (Mistral and LLaMA 3.1) in all settings except asking about Ukraine in English, where LLaMA 3.1 is better.

Breakdown per question category (Table~\ref{tab:appendix_breakdown} in the Appendix) shows that models score the highest on geography, followed by history and culture.

\paragraph{Visual QA.}

As shown in Table~\ref{tab:baselines}, the results of the visual subtask are much lower. Whereas in textual QA, the accuracy of the best-scoring model consistently exceeds 40\% in the strictest setting, with visual QA, it does not exceed 30\% in any of the languages.

Results of all vision models are presented in Table~\ref{tab:detailed_baselines_visual}.
By a large margin, the best-scoring model is LLaMA 4 Scout, followed by Gemma 3 and LLaMA 3.2. Idefics and Maya score very poorly, with Maya achieving near-zero performance across all languages and conditions. The visual QA results show even greater regional variation than textual QA. For instance, Llama 4 Scout performs notably better for Ukraine compared to Czechia or Slovakia.
Commercial models are by a large margin better than open-weight models. Note, however, that whilst in the textual scenario, the performance of LLaMA 3.3 was comparable to the lightweight closed-source models (Gemini 2.5 Flash-Lite, GPT-5-mini), in case of visual QA, we did not have access to a open-weight vision-language model of comparable size.
The question category breakdown  (Table~\ref{tab:appendix_breakdown} in the Appendix) does not show consistent patterns across countries and languages.

When we consider the visual content of the images (see Table~\ref{tab:appendix_visual_breakdown}), we observe a difference in questions about images with faces in Czechia and Slovakia on the one side, and Ukraine on the other. Whereas for Czechia and Slovakia, the models often refuse to answer (even though people in the images are public figures), in Ukrainian, images with faces are the best-scoring image category. Questions about images with cities score relatively well across languages and locations, followed by pictures of buildings. 
The presence of text increases the likelihood that the answer will be correct, even in cases where the text is not part of the answer but still provides some clues. Questions with color photos receive, on average, better answers than questions about black-and-white images. However, the confounding factor might be that questions about black-and-white images ask about older facts.

\begin{table*}[t]

\newcommand{\bglng}[1]{%
    \tikz[baseline=(text.base)]{%
        \node[draw=Gray, rounded corners=2pt, fill=Gray!40, inner sep=3pt, outer sep=0pt] (text) {#1};%
    }%
}
\newcommand{\bglngsm}[1]{%
    \tikz[baseline=(text.base)]{%
        \node[draw=Gray, rounded corners=2pt, fill=Gray!40, inner sep=1pt, outer sep=0pt] (text) {#1};%
    }%
}

\footnotesize\centering
\setlength{\tabcolsep}{4pt}

\begin{tabular}{ll cccc cccc cccc}
\toprule
& & \multicolumn{4}{c}{Czechia} & \multicolumn{4}{c}{Slovakia} & \multicolumn{4}{c}{Ukraine} \\ \cmidrule(lr){3-6} \cmidrule(lr){7-10} \cmidrule(lr){11-14}
& & \bglng{cs} & sk & uk & en & cs & \bglng{sk} & uk & en & cs & sk & \bglng{uk} & en \\ \midrule
\multirow{5}{*}{\rotatebox{90}{Textual}}
& EuroLLM-9B-Instruct & \bf\XlingTextualCZ{35.5} & \XlingTextualCZ{29.8} & \XlingTextualCZ{8.3} & \XlingTextualCZ{31.7}   & \XlingTextualSK{29.0} & \XlingTextualSK{30.2} & \XlingTextualSK{9.7} & \bf\XlingTextualSK{32.9}   & \XlingTextualUA{17.4} & \XlingTextualUA{15.8} & \XlingTextualUA{26.8} & \bf\XlingTextualUA{29.1}   \\
& Llama-3.1-8B-Instruct & \bf\XlingTextualCZ{28.5} & \XlingTextualCZ{17.0} & \XlingTextualCZ{8.9} & \XlingTextualCZ{31.3}   & \XlingTextualSK{21.5} & \XlingTextualSK{18.3} & \XlingTextualSK{7.9} & \bf\XlingTextualSK{25.2}   & \XlingTextualUA{18.7} & \XlingTextualUA{15.1} & \XlingTextualUA{23.4} & \bf\XlingTextualUA{34.8}   \\
& Llama-3.3-70B-Instruct & \bf\XlingTextualCZ{57.5} & \XlingTextualCZ{52.3} & \XlingTextualCZ{30.0} & \XlingTextualCZ{56.2}   & \XlingTextualSK{43.6} & \XlingTextualSK{43.8} & \XlingTextualSK{24.9} & \bf\XlingTextualSK{47.1}   & \XlingTextualUA{40.8} & \XlingTextualUA{39.0} & \bf\XlingTextualUA{54.5} & \XlingTextualUA{54.3}   \\
& Llama-4-Scout-17B-16E-Instruct & \bf\XlingTextualCZ{38.3} & \XlingTextualCZ{33.2} & \XlingTextualCZ{21.5} & \XlingTextualCZ{38.1}   & \XlingTextualSK{28.6} & \XlingTextualSK{32.9} & \XlingTextualSK{18.7} & \bf\XlingTextualSK{33.5}   & \XlingTextualUA{31.7} & \XlingTextualUA{34.8} & \bf\XlingTextualUA{47.3} & \XlingTextualUA{46.8}   \\
& Mistral-7B-Instruct-v0.3 & \XlingTextualCZ{22.6} & \XlingTextualCZ{11.9} & \XlingTextualCZ{11.7} & \bf\XlingTextualCZ{26.8}   & \XlingTextualSK{17.2} & \XlingTextualSK{13.6} & \XlingTextualSK{8.9} & \bf\XlingTextualSK{22.7}   & \XlingTextualUA{17.4} & \XlingTextualUA{12.2} & \XlingTextualUA{21.8} & \bf\XlingTextualUA{32.5}   \\
\midrule
\multirow{5}{*}{\rotatebox{90}{Visual}}
& gemma3 & \XlingVisualCZ{10.6} & \XlingVisualCZ{7.5} & \XlingVisualCZ{5.8} & \bf\XlingVisualCZ{12.8}   & \XlingVisualSK{12.7} & \bf\XlingVisualSK{13.6} & \XlingVisualSK{7.6} & \XlingVisualSK{11.9}   & \XlingVisualUA{13.7} & \XlingVisualUA{12.7} & \bf\XlingVisualUA{22.1} & \XlingVisualUA{21.1}   \\
& idefics & \bf\XlingVisualCZ{7.5} & \XlingVisualCZ{3.5} & \XlingVisualCZ{4.4} & \XlingVisualCZ{6.6}   & \XlingVisualSK{6.8} & \XlingVisualSK{7.6} & \XlingVisualSK{8.5} & \bf\XlingVisualSK{10.2}   & \XlingVisualUA{7.8} & \XlingVisualUA{4.9} & \XlingVisualUA{8.3} & \bf\XlingVisualUA{10.3}   \\
& Llama-3.2-11B-Vision-Instruct & \XlingVisualCZ{10.2} & \XlingVisualCZ{7.5} & \XlingVisualCZ{8.4} & \bf\XlingVisualCZ{13.7}   & \bf\XlingVisualSK{11.9} & \XlingVisualSK{9.3} & \XlingVisualSK{11.0} & \XlingVisualSK{11.0}   & \XlingVisualUA{11.8} & \XlingVisualUA{9.3} & \XlingVisualUA{15.2} & \bf\XlingVisualUA{19.1}   \\
& Llama-4-Scout-17B-16E-Instruct & \XlingVisualCZ{15.0} & \XlingVisualCZ{11.1} & \XlingVisualCZ{4.9} & \bf\XlingVisualCZ{15.9}   & \XlingVisualSK{16.1} & \bf\XlingVisualSK{18.6} & \XlingVisualSK{7.6} & \XlingVisualSK{16.1}   & \XlingVisualUA{16.2} & \XlingVisualUA{14.2} & \XlingVisualUA{25.0} & \bf\XlingVisualUA{29.9}   \\
& maya & \bf\XlingVisualCZ{1.8} & \XlingVisualCZ{0.0} & \XlingVisualCZ{0.4} & \bf\XlingVisualCZ{1.8}   & \XlingVisualSK{0.0} & \XlingVisualSK{0.0} & \XlingVisualSK{0.0} & \XlingVisualSK{0.0}   & \XlingVisualUA{2.9} & \XlingVisualUA{1.5} & \XlingVisualUA{2.9} & \bf\XlingVisualUA{5.9}   \\

\bottomrule\end{tabular}
\caption{Cross-lingual comparison of models: We report the accuracy by LLaMA 3.3 70B across translations of the questions and answers into other languages from the dev subset (with the local language \bglngsm{highlighted}). Note that the accuracies are only comparable within the country-specific blocks.}\label{tab:xlingcompare}

\end{table*}
\subsection{Metrics and Human Judgment}\label{ssec:hum_judgment}

We examine the correlation between automatic evaluation metrics and human judgments by comparing system-level (Table~\ref{tab:system_level}) and answer-level (Table~\ref{tab:answer_level}) performance.

\paragraph{System-level correlation.}

For textual QA, we find very high correlations between most metrics and human judgment. Traditional string-overlap metrics like BLEU, chrF, and ROUGE-L achieve correlations above 0.85 with human correctness judgments. This strong performance likely stems from the high proportion of named entities in the answers, where exact string matches are more meaningful than in other generation tasks like machine translation or summarization.
LLM-based evaluation metrics show even stronger correlations, with LLaMA 3.3 70B achieving near-perfect system-level correlation ($r > 0.95$) across most conditions. M-Prometheus, specifically designed for multilingual evaluation, performs similarly well. Interestingly, BERTScore shows lower correlations than simpler string-overlap metrics.

This changes considerably for visual QA, where correlations drop substantially across all metrics except for LLM as a judge. String-overlap metrics maintain reasonable performance (correlations around 0.6--0.8), but BERTScore shows negative correlations in some cases.

\paragraph{Answer-level correlation.}

At the answer level, correlations are generally lower but follow similar patterns. LLM-based metrics consistently outperform other metrics, achieving correlations around 0.6--0.8 with human judgments for both modalities. The difference between system-level and answer-level correlations shows that, while automatic metrics can reliably rank systems on average, they are less effective at judging individual answer quality.

\paragraph{LLM as a judge.}

We further analyzed the LLaMA 3.3 70B as a judge. Table~\ref{tab:confusion} shows the confusion matrices. The number of false positives and false negatives is relatively balanced. When exploring the data, we noticed that questions that get misclassified by the judge repeat across models, both as false positive and false negatives. However, we were not able to find a pattern explaining what they have in common E.g., in Czech, singer Karel Gott, football club Baník Ostrava or the Estates Theatre in Prague systematically confused the judge.

\vspace{0.8\baselineskip}
\noindent
Our findings reveal that simple string-overlap metrics perform surprisingly well for textual QA, likely because named entities are more prevalent. However, LLM-based evaluation seems to be the most reliable approach across all conditions, particularly for the more challenging visual QA task.

\subsection{Cross-Lingual Differences}\label{ssec:xling_diff}

The results of cross-lingual comparison (Table~\ref{tab:xlingcompare}) show substantial performance differences depending on the language of the question.

For textual QA, models tend to perform well when answering questions in the local language. However, the performance gaps vary across models and countries.
For questions about Czechia, the largest model (LLaMA 3.3 70B) maintains stable performance for Czech, Slovak, and English. Smaller models exhibit greater variability, often performing better in English than in the translated languages.
When questions about Slovakia are translated into Czech, the performance remains roughly the same, whilst when translated into Ukrainian, the performance deteriorates. Finally, answering in English yields better results than answering in Slovak itself.
For questions about Ukraine, the cross-lingual performance pattern differs: models answer questions about Czechia and Slovakia with comparable accuracy in Czech and Slovak, but performance decreases when these questions are asked in Ukrainian. Likewise, when answering questions about Ukraine, models perform poorly if the questions are translated into Czech or Slovak.

Visual QA shows even bigger variations. However, relatively low accuracy scores might affect the reliability of the statistics. Whereas Czech is generally the best-performing language for questions about Czechia, it is better to query the models in English for questions about Slovakia and Ukraine. Surprisingly, for questions about Slovakia, the LLaMA 3 models worked better in Czech than in Slovak.

\begin{table}
\centering\footnotesize
    \setlength{\tabcolsep}{3.6pt}

\begin{tabular}{l cc cc cc}
    \toprule
    \multirow{2}{*}{Model} & \multicolumn{2}{c}{CZ} & \multicolumn{2}{c}{SK} & \multicolumn{2}{c}{UA} \\ \cmidrule(lr){2-3} \cmidrule(lr){4-5} \cmidrule(lr){6-7}
    & cs & en & sk & en & uk & en \\ \midrule
    EuroLLM 9B Ins. & 1.6 & 2.4 & 0.7 & 0.7 & 1.9 & 1.5 \\
    Llama 3.1 8B Ins. & 1.9 & 2.3 & 1.3 & 3.2 & 1.4 & 2.2 \\
    Llama 3.3 70B Ins. & 1.0 & 1.6 & 1.4 & 1.8 & 1.0 & 0.6 \\
    Llama 4 Scout Ins. & 1.0 & 1.3 & 1.2 & 0.7 & 1.5 & 2.7 \\
    Mistral 7B Ins. v0.3 & 1.3 & 0.4 & 0.7 & 0.7 & 1.0 & 0.7 \\
    \bottomrule
\end{tabular}

\caption{Standard deviation of the LLM-as-a-judge score when running the models with different prompts.}\label{tab:prompt_deviation}

\end{table}

\subsection{Prompt Variation} %
\label{ssec:promptvars}

To verify that the presented results are not an artifact of the prompts we selected, we conduct additional experiments to assess the dataset's robustness to prompt variations.
For each location, we tested additional 4 prompts in the local language and 4 in English, varying in wording and style (see Appendix~\ref{app:prompts}). 

Table~\ref{tab:prompt_deviation} shows standard deviations of scores given by LLM as a judge, evaluated on the development set. In most cases, the deviation is around one percentage point, which is smaller than the differences between models. The responses to more colloquial prompts differ in style, but rarely in the factual content. The average chrF and LLM-as-a-judge scores are in Table~\ref{tab:prompt_variation} in the Appendix.

\begin{table}
\centering\footnotesize
    \setlength{\tabcolsep}{3.6pt}

\begin{tabular}{l cc cc cc}
    \toprule
    \multirow{2}{*}{Model} & \multicolumn{2}{c}{CZ} & \multicolumn{2}{c}{SK} & \multicolumn{2}{c}{UA} \\ \cmidrule(lr){2-3} \cmidrule(lr){4-5} \cmidrule(lr){6-7}
    & cs & en & sk & en & uk & en \\ \midrule
    EuroLLM 9B Ins.      & \XlingTextualCZ{37.7} & \XlingTextualCZ{33.0}
                         & \XlingTextualSK{68.5} & \XlingTextualSK{60.2}
                         & \XlingTextualUA{27.3} & \XlingTextualUA{19.2}\\
                         
    Llama 3.1 8B Ins.    & \XlingTextualCZ{52.3} & \XlingTextualCZ{47.0}
                         & \XlingTextualSK{69.8} & \XlingTextualSK{47.0}
                         & \XlingTextualUA{44.2} & \XlingTextualUA{33.7} \\
                         
    Llama 3.3 70B Ins.   & \XlingTextualCZ{65.9} & \XlingTextualCZ{59.8}
                         & \XlingTextualSK{76.9} & \XlingTextualSK{73.4}
                         & \XlingTextualUA{58.7} & \XlingTextualUA{51.2} \\
    
    Mistral 7B Ins. v0.3 & \XlingTextualCZ{50.4} & \XlingTextualCZ{47.2}
                         & \XlingTextualSK{71.2} & \XlingTextualSK{63.3}
                         & \XlingTextualUA{46.0} & \XlingTextualUA{33.5} \\
    
    \bottomrule
\end{tabular}

\caption{Accuracy of retrieval-augmented generation using local Wikipedia evaluated with LLaMA 3.3 70B as a judge on the development set.}\label{tab:rag}

\end{table}

\subsection{Retrieval Augmented Generation}%
\label{ssec:ragresults}

The results are presented in Table~\ref{tab:rag}. As expected, having access to the retrieved context leads to superior performance in terms of LLM as a judge score.
LLaMA 3.3 70B is still the best performing model. The results of the Mistral model are especially noteworthy, as its performance increased dramatically in the local language settings.
EuroLLM seem not to benefit from the additional context, which suggests weaker long context handling abilities.

All models perform better with RAG in all setups, except when asking about Ukraine in English.
Upon manual inspection of the outputs, we noticed that the low scores in English questions about Ukraine are caused by the models often answering in Ukrainian, despite the instructions.
The biggest scores were achieved on Slovak questions. This is likely caused by smaller size of the Slovak Wikipedia and shorter article lengths, which makes the retrieval more accurate and the prompts shorter.

\section{Discussion and Future Work}\label{sec:discussion}

Our experiments show that current evaluation metrics work reasonably well for open-ended regional QA. This finding suggests that the focus on multiple-choice QA in many existing datasets may be unnecessarily restrictive. The high proportion of named entities in factual answers makes string-overlap metrics more effective than in other generation tasks.

The dataset can serve multiple research purposes beyond basic QA evaluation. Since all information was available on Wikipedia as of late 2024, researchers can use it to evaluate RAG systems or cross-lingual RAG approaches. This provides a controlled setting where the knowledge source is known and accessible.

The question of what knowledge should be built into models versus retrieved from external sources remains open. Regional knowledge presents an interesting test case for this trade-off. While one could argue that such specific information should be left to external knowledge sources, we believe that broader factual knowledge generally improves a model's ability to search for and contextualize additional information. Models that perform better on our dataset are likely to provide more culturally appropriate responses across different regions.

CUS-QA is designed as an evaluation benchmark to measure model performance on regional knowledge. The dataset supports assessment across both textual QA and visual QA, with visual QA proving significantly more difficult.
The dataset also enables studies of cross-lingual consistency that allow researchers to examine how well models transfer knowledge across related languages.

To make the benchmark more accessible, we host it on Codabench \cite{codabench}, where users can test their models. Participants can download the full dev set and the test set without the reference answers. The platform will then score the prediction on the full test set, with results broken by language and modality. In this way, participants can score partial submissions and focus only on a subset of modalities and languages.
On Codabench, we provide all the metrics with the exception of LLM as a judge with M-Prometheus due to hardware limitations.

Our human evaluation provides a resource for developing and validating automatic metrics for open-ended QA. Since human-labeled datasets for this task are rare, our annotations offer a benchmark for future research on reference-based evaluation and metric reliability.

Finally, the dataset can serve as a seed for automatically generating additional training data from Wikipedia or other resources. Researchers can extend our approach to create larger-scale datasets covering more regions and languages, or use similar collection methods for other knowledge domains.

\begin{table*}[t]
\newcommand{\cmark}{\ding{51}}  
\newcommand{\xmark}{\textcolor{Gray!60}{\ding{55}}}  
\newcommand{\pmark}{\tikz[baseline=-0.5ex]\draw[line width=1pt,dotted] (0,0) circle (0.7ex);}  
\centering\footnotesize
\setlength{\tabcolsep}{5pt}
\begin{tabular}{ll ccccccccc}
\toprule
\multirow{2}{*}{Dataset name} & \multirow{2}{*}{Citation} & \multicolumn{3}{c}{Languages} & \multicolumn{2}{c}{Modality} & \multirow{2}{*}{\begin{minipage}{1cm}\centering Regional questions\end{minipage}} & \multirow{2}{*}{\begin{minipage}{1cm}\centering Human eval.\end{minipage}} & \multirow{2}{*}{\begin{minipage}{2cm}\centering Size in \\ cs+sk+uk\end{minipage}} \\ \cmidrule(lr){3-5} \cmidrule(lr){6-7}
& & cs & sk & uk & Text & Vis. & & &  \\
\midrule
MultiLoKo & \citet{hupkes2025multiloko} & \cmark & \xmark & \xmark & \cmark & \xmark & \cmark & \xmark & 500 \\
INCLUDE & \citet{romanou2024include} & \cmark & \cmark & \cmark & \cmark & \xmark & \xmark & \xmark & 50+31+182 \\
Global MLLU & \citet{singh-etal-2025-global} & \cmark & \xmark & \cmark & \cmark & \xmark & \xmark & \xmark & 14k \\
BenCzechMark & \citet{fajcik2025benczechmark} & \cmark & \xmark & \xmark & \cmark & \xmark & \pmark & \xmark &  \\
ZNO & \citet{paniv-etal-2025-benchmarking} & \xmark & \xmark & \cmark & \cmark & \xmark & \pmark & \xmark & 3.7k \\
ALM-bench & \citet{vayani2025all} & \cmark & \cmark & \cmark & \cmark & \cmark & \pmark & \xmark & 269+128+269 \\
WorldCuisines & \citet{winata-etal-2025-worldcuisines} & \cmark & \xmark & \xmark & \cmark & \cmark & \xmark & \xmark & 1.5k \\

\midrule

CUS-QA &  & \cmark & \cmark & \cmark & \cmark & \cmark & \cmark & \cmark & 1536+1210+1158 \\
\bottomrule
\end{tabular}

\caption{Comparison of CUS-QA with other existing QA datasets for Czech, Slovak and Ukrainian. The symbol \pmark{} in the Regional question column indicate that the dataset contains some regional questions, but it was not the focus of the dataset.}\label{tab:datasets}
\end{table*}

\section{Related Work}\label{sec:related_work}

We review existing datasets for multilingual question answering (\S~\ref{ssec:txtqa}), visual question answering (\S~\ref{ssec:vqa}), dataset targeting our languages of interest (\S~\ref{ssec:csskuk}), and evaluation metrics (\S~\ref{ssec:metrics}).

\subsection{Textual Question Answering}\label{ssec:txtqa}

With the rise of decoder-based generative models, multiple-choice QA has become a common evaluation method. The most frequently used English MMLU dataset \citep{hendrycks2021measuring} covers 57 diverse topics but is English-only and assumes US-centric knowledge in some areas.

Several datasets extend QA evaluation to other languages with regional content. \citet{etxaniz2024bertaqa} introduce parallel trivia questions in English and Basque, with a subset on Basque culture. KazQAD \citep{yeshpanov-etal-2024-kazqad} combines Kazakh local knowledge with general subjects from school exams. \citet{kostiuk2024towards} compile a Lithuanian history dataset, showing that prompting in Lithuanian improves performance. \citet{etori2025lag} localize MMLU into Latvian and Giriama.

Larger-scale multilingual efforts include INCLUDE \citep{romanou2024include}, which collects multiple-choice questions from local exam sources in 44 languages, and Global MMLU \citep{singh2024global}, which covers 42 languages with both culture-agnostic and culture-specific questions. Food-focused datasets use cuisine as a cultural proxy: \citet{zhou2024mapo} presents questions on local ingredients, while \citet{lavrouk2025foundation} covers dishes from former Soviet states, revealing that LLMs often confuse post-Soviet nations.

Cross-lingual performance gaps have been documented in several studies. \citet{rohera2025better} find that LLMs often perform better in English even for questions from Indic contexts. \citet{hupkes2025multiloko} show more nuanced results with MultiLoKo, a 31-language benchmark with 500 locally-sourced questions per language. \citet{goldman2025ecklektic} present ECLeKTic, which uses Wikipedia article presence across 12 languages to identify language-specific facts and evaluate cross-lingual knowledge transfer, showing that current models struggle to share knowledge across languages even when they can answer in the source language.

\subsection{Visual Question Answering}\label{ssec:vqa}

Initial formulations of Visual Question Answering (VQA; \citealp{antol2015visual,hudson2019gqa}) assumed a closed output vocabulary to evaluate object understanding, an approach carried into multilingual extensions \citep{pfeiffer-etal-2022-xgqa}.

ALM-Bench \citep{vayani2025all} is most similar to our work, covering cultural elements from 73 countries in both local languages and English, and including several question types: short and long open-ended, binary, and multiple-choice. CVQA \citep{romero2024cvqaculturallydiversemultilingualvisual} covers 30 countries, but it uses a multiple-choice format and does not overlap with our target regions. \citet{nayak-etal-2024-benchmarking} introduce an English-only benchmark for open-ended VQA on traditions, food, and clothing. \citet{wang2025travelinglanguagesbenchmarkingcrosslingual} propose a multiple-choice VQA set on tourism.
Other multimodal benchmarks incorporate exam questions across languages \citep{zhang2023m3exammultilingualmultimodalmultilevel,das-etal-2024-exams}, but focus on visual elements, such as graph comprehension, rather than cultural knowledge. 

\subsection{QA in Czech, Slovak, and Ukrainian}\label{ssec:csskuk}

Table~\ref{tab:datasets} summarizes existing datasets covering our target languages.

MultiLoKo \citep{hupkes2025multiloko} uses a similar Wikipedia-based methodology but covers only Czech with 500 questions. INCLUDE \citep{romanou2024include} provides school exam questions in all three languages, but with a limited size and inconsistent domains. Global MMLU \citep{singh-etal-2025-global} does not include Slovak and relies mostly on unverified machine translation for Czech and Ukrainian.

For single-language resources, BenCzechMark \citep{fajcik2025benczechmark} aggregates Czech datasets, including school exams with some regional content. 
The ZNO dataset \citep{paniv-etal-2025-benchmarking} comprises 3.7k Ukrainian national exam questions covering also regional history and literature.

Two datasets include visual QA: ALM-bench \citep{vayani2025all} covers all three languages with culturally-focused questions that were machine-translated and post-edited. WorldCuisines \citep{winata-etal-2025-worldcuisines} includes Czech but focuses narrowly on cuisine with minimal coverage of our target regions.

CUS-QA provides larger coverage, both modalities, explicit regional focus, and human evaluation annotations.

\subsection{Evaluation Metrics}\label{ssec:metrics}

Traditional metrics measure textual overlap using word or character n-grams (e.g., BLEU: \citealp{papineni-etal-2002-bleu}; ROUGE: \citealp{lin-2004-rouge}; chrF: \citealp{popovic-2015-chrf}). More recent approaches leverage embedding similarity (e.g., BERTScore: \citealp{zhang2020bertscore}, BLEURT: \citealp{sellam-etal-2020-bleurt}) or fully trained metrics like COMET \citep{rei-etal-2020-comet} that are targeted at machine translation.

Recently, LLMs have often been used as judges, either directly assessing the quality \citep{zheng2023judge} or as part of more structured pipelines, such as FactScore \citep{min-etal-2023-factscore}, which first breaks the generated text into atomic facts and evaluates them individually.

The standard approach to validating evaluation metrics is to measure their correlation with human judgments, a practice best established in machine translation \citep[e.g.,][]{freitag-etal-2024-llms}. However, human-labeled datasets for other tasks are limited, with exceptions like image captioning (Flickr8k-Expert \citealp{hodosh2013framing}; Nebula \citealp{matsuda2024nebula}; Polaris \citealp{wada2024polos}), text summarization (ROSE: \citealp{liu-etal-2023-revisiting}), and MOCHA for reading comprehension \citep{chen-etal-2020-mocha}, mostly in English.

\section{Conclusions}\label{sec:conclusions}

We introduce CUS-QA, a regional knowledge dataset covering textual and visual question answering in Czech, Slovak, and Ukrainian. The dataset contains manually curated questions and answers from native speakers, grounded in Wikipedia and focused on local knowledge that is well-known within each country but largely unknown outside it.

Our baseline experiments reveal significant gaps in regional knowledge capabilities of the current LLMs. The best textual QA model achieves above 40\% accuracy, while visual QA is much more challenging with accuracy below 30\%. We observe substantial cross-lingual inconsistencies, where models sometimes perform better answering regional questions in English rather than the local language.

Through human evaluation, we find that LLM-based metrics correlate well with human judgment for this task, while traditional string-overlap metrics perform surprisingly well due to the high number of named entities in the dataset, especially in the textual QA part. These findings suggest that open-ended evaluation of factual QA might be more feasible than previously thought.

The dataset addresses two key research needs: evaluating regional knowledge in LLMs, including cross-lingual generation consistency, and validating automatic evaluation metrics for open-ended QA. As an evaluation benchmark, CUS-QA enables researchers to measure model capabilities on culturally specific knowledge without training on the test distribution. We release both the dataset and human evaluation results to support future research in these areas.

\section*{Limitations}

Our dataset has several limitations. First, the dataset size is relatively small compared to major benchmarks. Second, all annotators were university students, which does not reflect the demography of the respective countries. Third, our focus is narrowed on three closely related Slavic languages.

The dataset is designed for evaluation only. With approximately 1,200--1,500 examples per language, it is too small for fine-tuning but appropriately sized for assessing model performance on regional knowledge.

The dataset reflects knowledge as of late 2024. Our reliance on Wikipedia as the grounding source introduces potential biases toward formal, institutional knowledge. Some overlap exists between Czech and Slovak questions due to shared history, potentially inflating cross-lingual performance comparisons.

The human evaluation was limited to five LLMs, which may not capture the full range of model outputs and error patterns that could emerge in future models. The binary evaluation criteria achieved good inter-annotator agreement but may not capture subtle quality differences between generated responses

Our evaluation focused solely on factual accuracy and excluded other important factors like cultural sensitivity, tone, and political awareness that could affect real-world performance.


%

\bibliography{anthology,custom}
\bibliographystyle{acl_natbib}


\onecolumn

\appendix

\section{Prompts}
\label{app:prompts}

We use all of the following prompts in testing the textual models, while for the visual only the 6th prompt was involved. All of them are followed by an image token (if the question is visual) and the question.
\begin{enumerate}
    \item You are an experienced trivia game contestant. Provide a short and correct answer!
    \item Odpověz pravdivě a krátce na následující otázku:
    \item Teraz odpovedajte na nasledujúcu otázku pravdivo a stručne:
    \item \cyr{Відповідайте на наступне питання правдиво і коротко:}
    \item Jsi zkušený účastník vědomostních soutěží. Odpovídej správně a stručně!
    \item You are an experienced trivia game contestant. Provide a short and correct answer in the same language as the question!
\end{enumerate}

\noindent For \textbf{LLM as a judge}, we used the following prompts:

\begin{itemize}

    \item System prompt: You are a judge in a trivia game and you are supposed to tell if the contestant answered the question correctly based on the question and the reference answer.

    \item User prompt: The question is: \{question\}  The contestant's answer is \{model\_answer\}. The correct answer is \{correct\_answer\}.  Answer 'Yes' or 'No' and nothing else.

\end{itemize}

\noindent For the \textbf{prompt variation study}, we used the following prompts:
\begin{itemize}[leftmargin=3em]

\item[(cs 1)] Jsi zkušený účastník vědomostních kvízů. Odpověz na násludující otázky správně a stručně.

\item[(cs 2)] Jsi odborník s neuvěřitelně širokým rozhledem a často vyhráváš populární vědomostní soutěže v rozhlase a televizi. Odpovídej na následující otázku stručně, správně a úplně.
\item[(cs 3)] Odpověz na následující otázku stručně a pravdivě!
\item[(cs 4)] Vědomostní soutěže jsou pro tebe úplně všechno. Roky jsi intenzivně studoval, abys teď uspěl ve vědomostních soutěžích. Všechno v tvém životě směřovalo k tomu, aby jsi teď bezchybně a stručně odpověděl na následující otázku.

\vspace{1em}

\item[(sk 1)] Si skúsený účastník kvízov. Odpovedz na nasledujúce otázky stručne!
\item[(sk 2)] Odpovedz na nasledujúce vedomostné otázky správne a čo najstručnejšie.
\item[(sk 3)] Si odborník na vedomostné kvízy. Správne a stručne zodpovedaj na otázky, ktoré dostaneš.
\item[(sk 4)] V škole si mal samé jednotky, mama ti pred spaním čítala encyklopédie, vyhral si Milionára a doma máš plnú policu fliaš s chľastom z pub kvízov. Zodpovedať nasledujúce otázky tak pre teba musí byť hračka. Správne a stručne.

\vspace{1em}

\item[(uk 1)] \cyr{Ви досвідчений учасник вікторин. Надайте коротку і правильну відповідь!}
\item[(uk 2)] \cyr{Ти досвідчений гравець у вікторини. Дай коротку та точну відповідь!}
\item[(uk 3)] \cyr{Ти експерт у вікторинах. Відповідай коротко і правильно!}
\item[(uk 4)] \cyr{Досвідчений учасник вікторин. Коротка правильна відповідь!}

\vspace{1em}

\item[(en 1)] You have extensive general knowledge. Answer the following question correctly and concisely.
\item[(en 2)] You're a trivia expert with an incredibly broad knowledge base and frequently win popular quiz competitions. Answer the following question briefly, correctly, and completely.
\item[(en 3)] Answer the following question concisely and truthfully!
\item[(en 4)] Trivia is your life. You've spent years devouring facts, competing in quiz leagues, and crushing every pub quiz in town. Your brain is a finely-tuned knowledge machine. Now prove it by answering the following question correctly and briefly.

\end{itemize}

\noindent We use the following prompts for \textbf{retrieval augmented generation}. 

\begin{itemize}
    \item You are an experienced trivia game contestant. Provide a short and correct answer to the question! Attached to the question is a snippet from the local Wikipedia. You may use the information in the context as you see fit. Do not mention the context snippet in your reply. Reply in the original language of the question.
\end{itemize}



\begin{table*}[t!]
\centering\footnotesize

\begin{tabular}{p{10cm}p{5cm}}

\toprule

\multicolumn{2}{c}{Czechia} \\ \midrule

Kdo byl jediným českým králem, který nepocházel z panovnické dynastie? & Jiří z Poděbrad. \\
\it Who was the only Czech king who did not come from a ruling dynasty? & \it George of Poděbrady \\
\midrule

V kterém roce proběhla bitva na Bílé hoře? & 1620 \\
\it In what year did the Battle of White Mountain take place? & \it 1620 \\
\midrule

Ve kterém městě sídlí Tatra? & V Kopřivnici. \\
\it In which city is the Tatra factory located? & \it In Kopřivnice. \\
\midrule

Jak se jmenuje zřícenina hradu založená Karlem IV. jižně od Starého Plzence, která je kulturní památkou? & Radyně \\
\it What is the name of the ruins of the castle founded by Charles IV south of Stary Plzenec, which is a cultural monument? & \it Radyně \\
\midrule

Který lihovar vyrábí OMG gin? & Žufánek \\
\it Which distillery produces OMFG gin? & \it Žufánek \\
\midrule

Která linka pražského metra je nejstarší? & Linka C \\
\it Which Prague metro line is the oldest? & \it The C line \\
\midrule

\multicolumn{2}{c}{Slovakia} \\

\midrule

Ktorá kapela naspievala album Nikdy nebolo lepšie? & Hex \\
\it Which band sang the album Never Been Better? & \it Hex \\
\midrule

Ako sa volá najvyšší vrch Belianskych Tatier? & Havran \\
\it What is the name of the highest peak of the Beliany Tatras? & \it Havran \\
\midrule

Kto založil Medzinárodný maratón mieru v Košiciach? & Vojtech Bukovský. \\
\it Who founded the International Peace Marathon in Košice? & \it Vojtech Bukovský. \\
\midrule

Ktorá televízna stanica vysiela šou s názvom Tvoja tvár znie povedome? & televízia Markíza \\
\it Which television station airs a show called Tvoja tvár znie povedome? & \it television Markíza \\
\midrule

Ako sa volal politický spor o názov Česko-Slovenska na začiatku roka 1990? & pomlčková vojna \\
\it What was the name of the political conflict regarding the name of Czechoslovakia at the beginning of the year 1990? & \it The Hyphen War \\
\midrule

Ktorá diaľnica prechádza nedaľeko Nového Mesta nad Váhom? & D1 \\
\it Which highway passes near Nové Mesto nad Váhom? & \it D1 \\

\midrule

\multicolumn{2}{c}{Ukraine} \\

\midrule

\cyr{На якій вулиці знаходиться Будинок з химерами у Харкові?} & \cyr{На вулиці Чернишевській} \\
\it What street is the House with Chimeras located on in Kharkiv? & \it On Chernyshevskaya Street \\
\midrule

\cyr{Який оператор мобільного зв'язку в 2015 викупив право надання послуг у компанії МТС?} & \cyr{Vodafone} \\
\it Which mobile operator bought the right to provide services from MTS in 2015? & \it Vodafone \\
\midrule

\cyr{Які кольори футбольного клубу Нива?} & \cyr{Жовтий і зелений} \\
\it What are the colors of Niva football club? & \it Yellow and green \\
\midrule

\cyr{Яку міжнародну премію зміг отримати Іван Данилович Низовий?} & \cyr{За найкращий музичний твір.} \\
\it What international award was Ivan Danylovych Nyzovyi able to receive? & \it For the best musical composition. \\
\midrule

\cyr{Як називається головна вулиця Києва?} & \cyr{Хрещатик} \\
\it What is the name of the main street in Kyiv? & \it Khreshchatyk \\
\midrule

\cyr{Хто очолював УНР до квітня 1918 року?} & \cyr{Михайло Грушевський} \\
\it Who led the UPR until April 1918? & \it Mykhailo Hrushevsky \\
\midrule
\end{tabular}

\caption{Examples of text-only questions and answers.}\label{tab:textual_extra}

\end{table*}

\begin{table*}

\centering\footnotesize

\begin{tabular}{l l p{6cm} p{4cm}}
\toprule
 & Image & Question & Answer \\ \midrule

\multirow{6}{*}{\rotatebox{90}{Czechia\hspace{58pt}}}
& \multirow{2}{*}{\includegraphics[width=2.5cm]{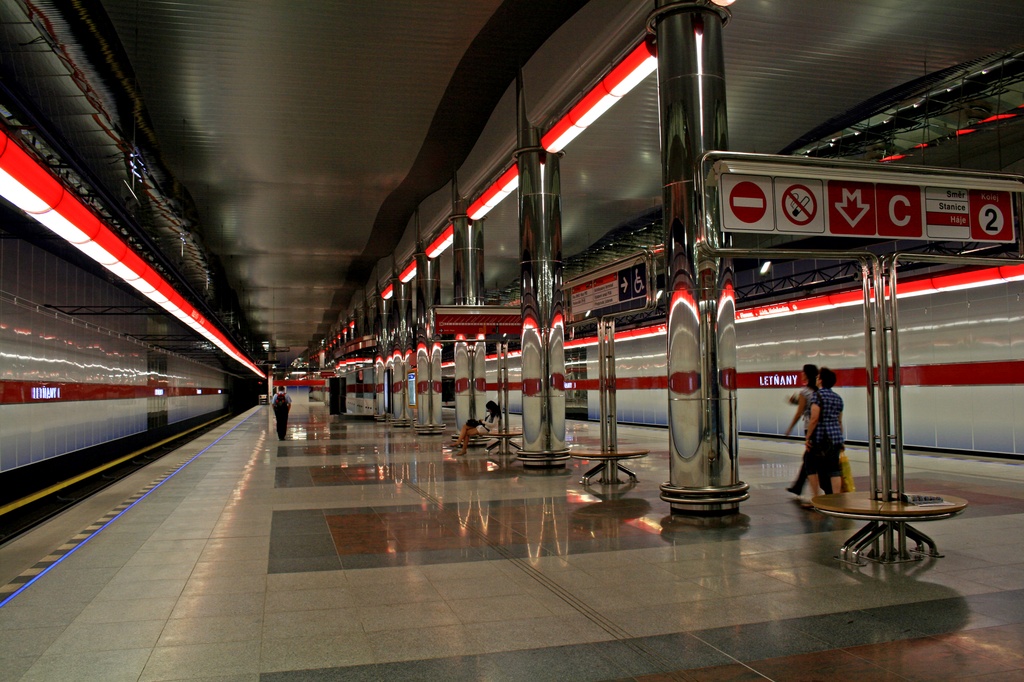}} &
Která stanice pražského metra je na obrázku? & Letňany \\
& & \it Which Prague metro station is in the picture? & \it Letňany \\[26pt]

& \multirow{2}{*}{\includegraphics[width=2.5cm]{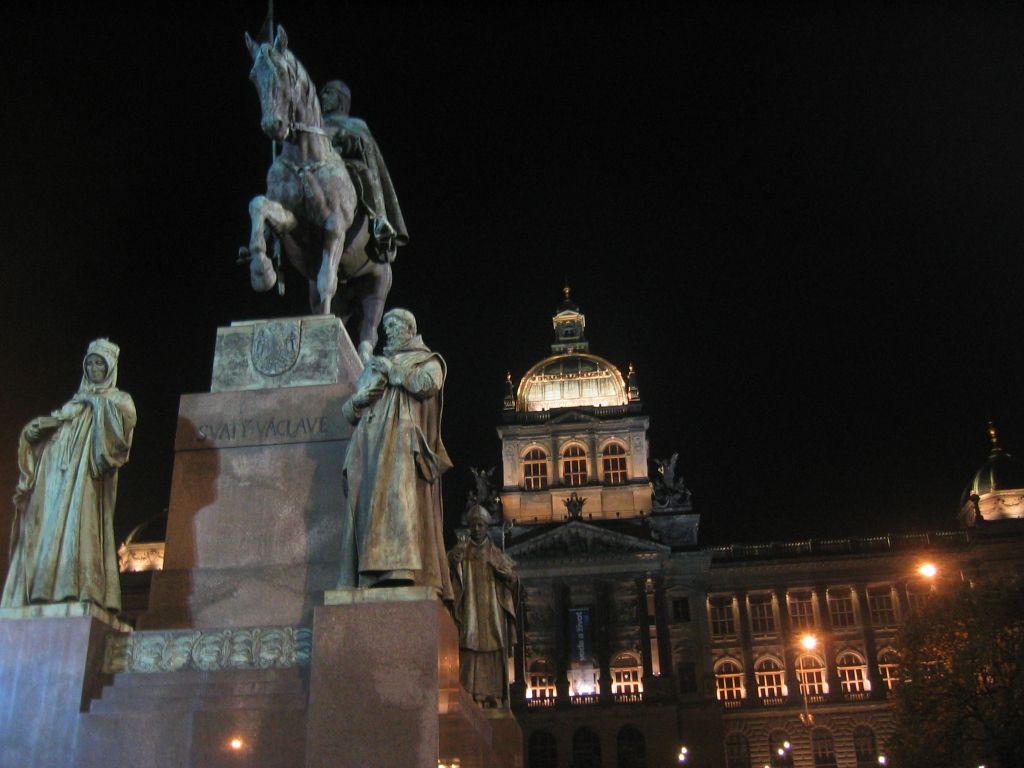}} &
Kdo je autorem této sochy? & Josef Václav Myslbek. \\
& & \it Who is the author of this statue? & \it Josef Václav Myslbek. \\ [32pt]

& \multirow{2}{*}{\includegraphics[width=2.5cm]{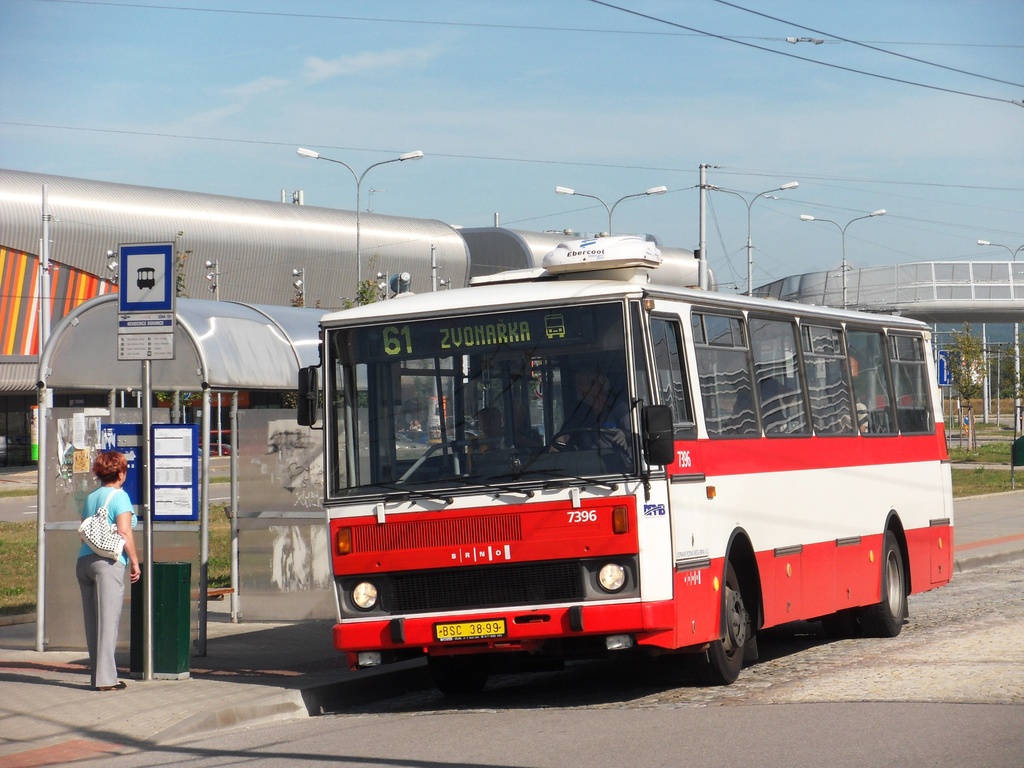}} &
Jaké značky je tento autobus? & Karosa. \\
& & \it What brand is this bus? & \it	Karosa. \\[32pt]

\midrule

\multirow{6}{*}{\rotatebox{90}{Slovakia\hspace{70pt}}}
& \multirow{2}{*}{\includegraphics[trim={20pt 250pt 18pt 240pt}, clip, %
width=2.5cm]{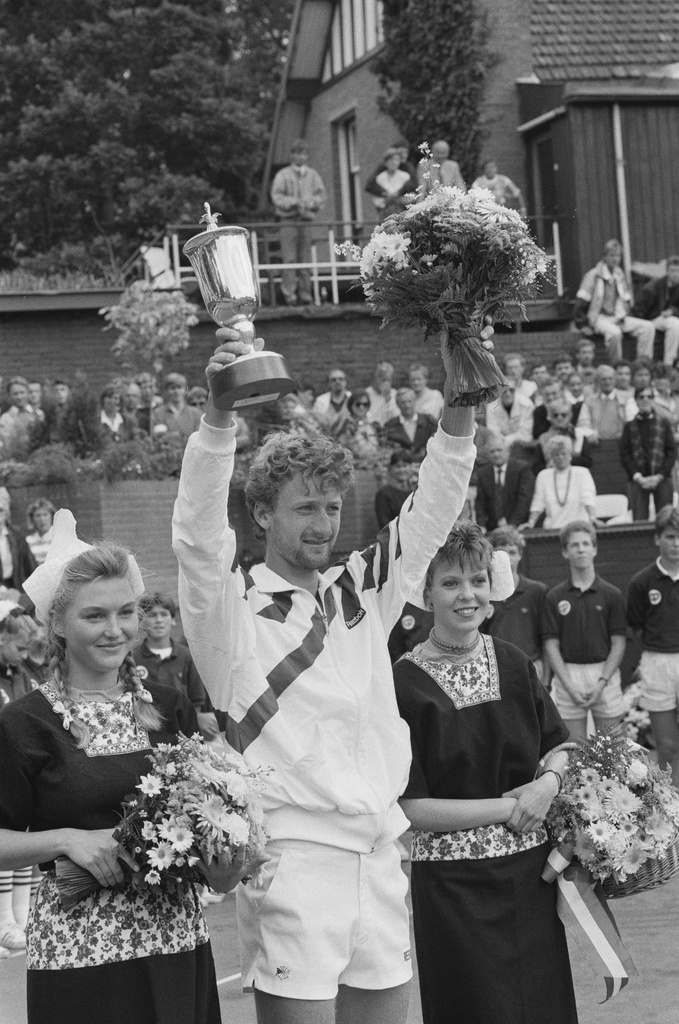}} & Ktorý bývalý slovenský tenista sa nachádza na fotke? & Miloslav Mečíř. \\
& & \it Which former Slovak tennis player is in the picture? & \it Miloslav Mečíř. \\ [27pt]

& \multirow{2}{*}{\includegraphics[width=2.5cm]{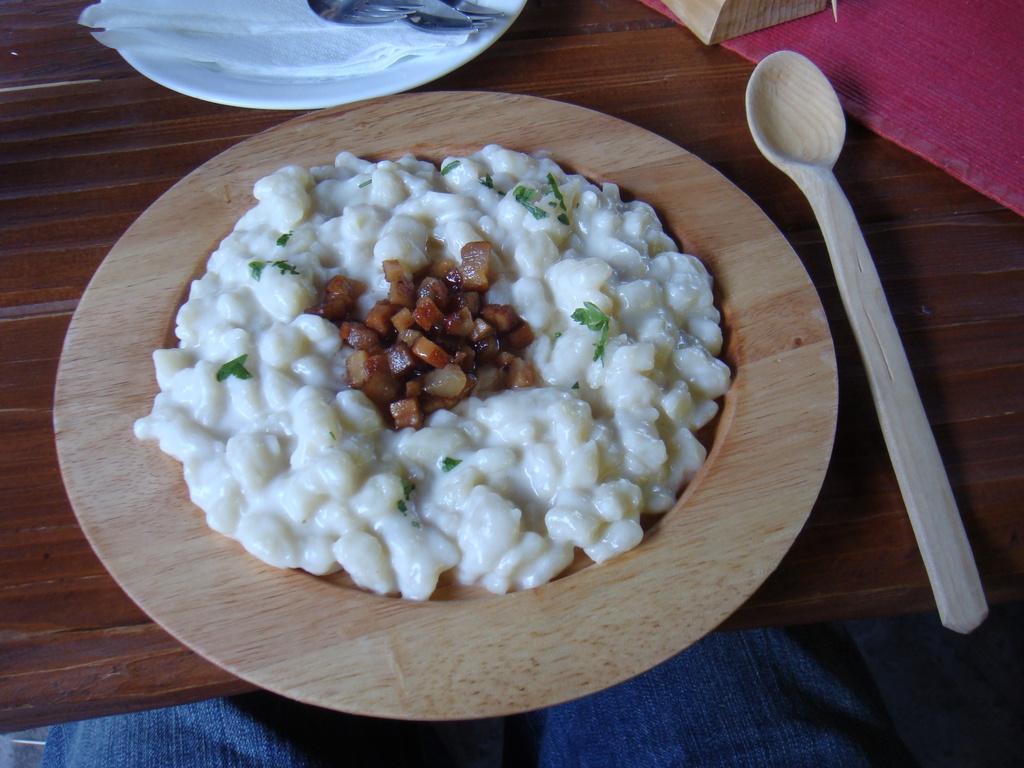}} &
Čo je na obrázku? & Bryndzové halušky. \\
& & \it What is in the picture? & \it Bryndzové halušky. \\ [32pt]

& \multirow{2}{*}{\includegraphics[width=2.5cm]{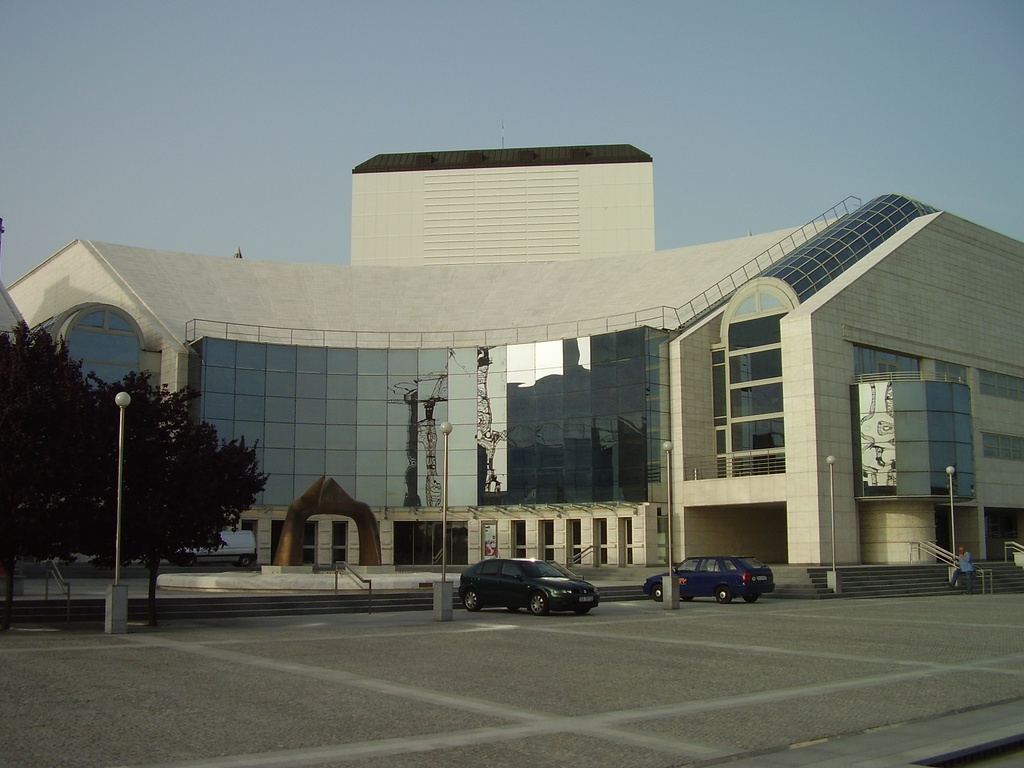}} &
Aká budova je na obrázku? & Nová budova Slovenského národného divadla \\
& & \it What building is in the picture? & \it New building of the Slovak National Theatre \\ [21pt]

\midrule

\multirow{6}{*}{\rotatebox{90}{Ukraine\hspace{70pt}}}
& \multirow{2}{*}{\includegraphics[trim={0cm 1.5cm 0cm 0cm}, clip, width=2.5cm]{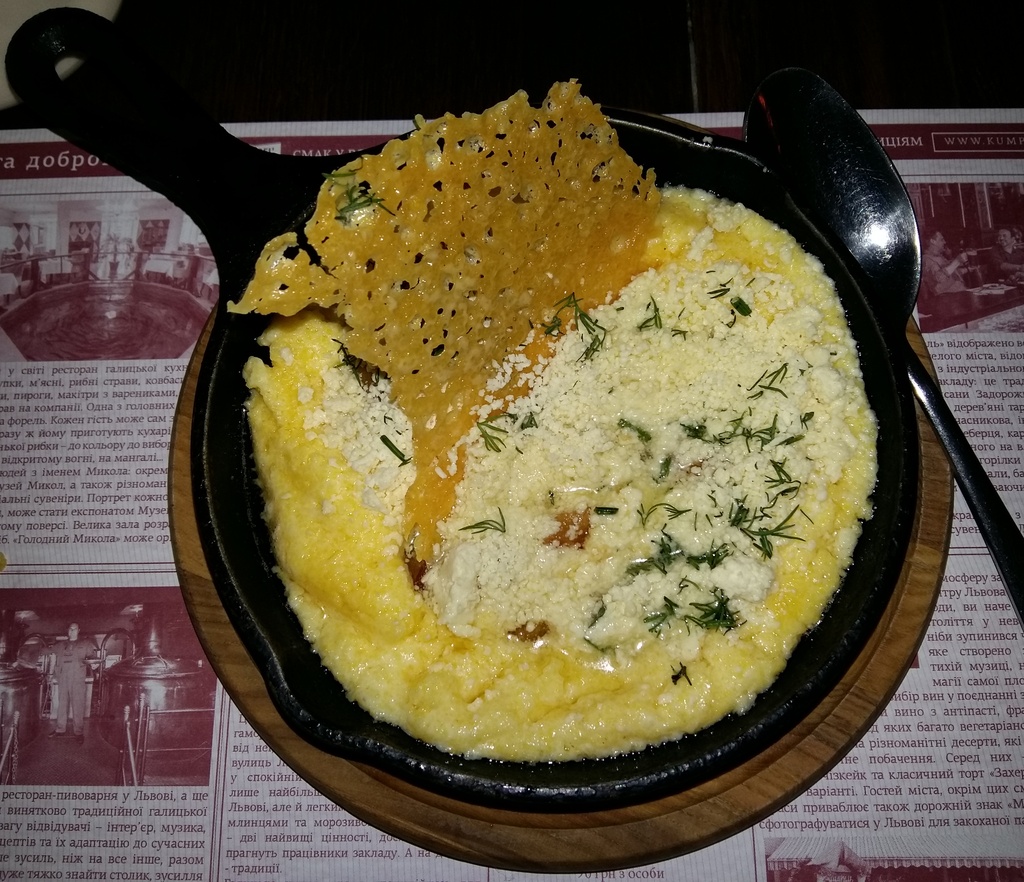}} & \cyr{Яку українську страву зображено на фото?} & \cyr{Банош} \\
& & \it What Ukrainian dish is depicted in the photo? &  Banosh \\ [26pt]

& \multirow{2}{*}{\includegraphics[width=2.5cm]{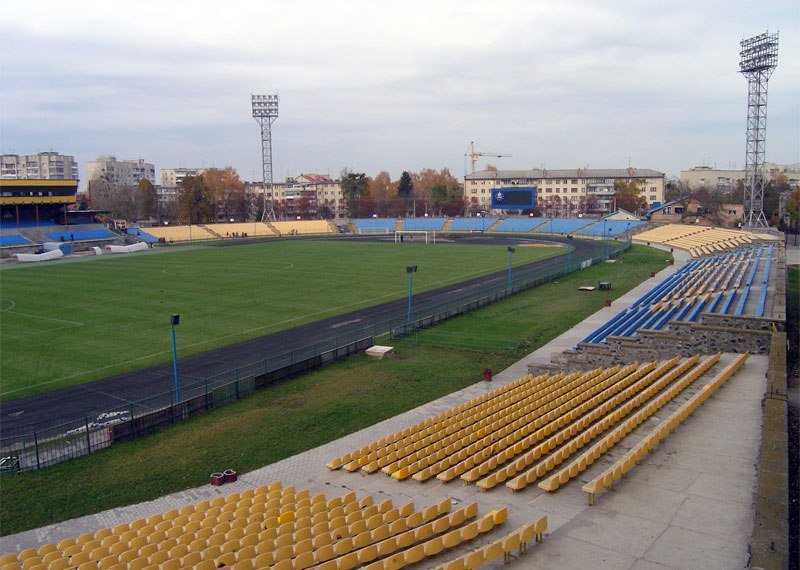}} &
\cyr{Як називається цей стадіон?} & \cyr{«Авангард»} \\
& & \it What is the name of this stadium? & ``Avangard'' \\ [29pt]

& \multirow{2}{*}{\includegraphics[width=2.5cm]{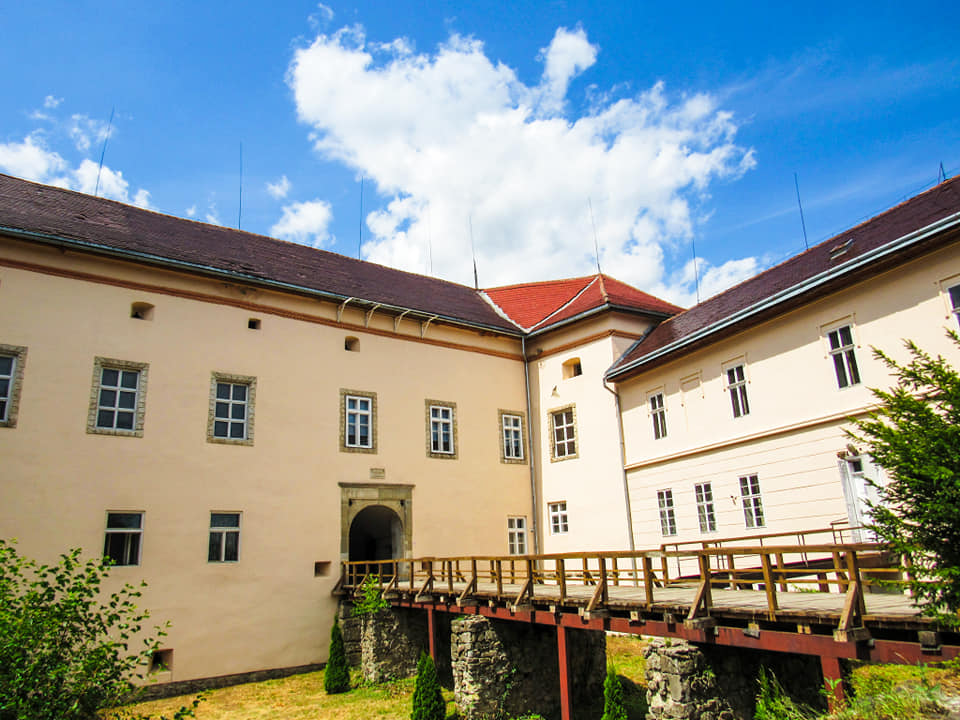}} &
\cyr{Де знаходиться зображений на картинці замок?} & \cyr{Ужгород} \\
& & \it Where is the castle in the picture located? & \it Uzhhorod \\[21pt]

\bottomrule
\end{tabular}

\caption{Examples of collected questions and answers from all regions, in both the local language and in English.}\label{tab:example_extra}
\end{table*}



\begin{table}


\footnotesize\centering
\setlength{\tabcolsep}{4.5pt}

\begin{tabular}{ll cccc ccccccc}
\toprule
\multicolumn{2}{l}{\multirow{2}{*}{Textual QA: Czechia}}
& \multicolumn{4}{c}{Manual evaluation}
& \multicolumn{7}{c}{Automatic evaluation}
\\
\cmidrule(lr){3-6}
\cmidrule(lr){7-13}
&
 & \begin{minipage}{6.6mm}\centering Correct \end{minipage}
 & \begin{minipage}{6.6mm}\centering + True    \end{minipage}
 & \begin{minipage}{6.6mm}\centering + \textls[-64]{Relevant}\end{minipage}
 & \begin{minipage}{6.6mm}\centering + \textls[-85]{Coherent}\end{minipage}

 & \begin{minipage}{6.2mm}\centering BL-EU\end{minipage}
 & \begin{minipage}{6.2mm}\centering chrF\end{minipage}
 & \begin{minipage}{6.2mm}\centering\textls[-85]{ROU-GE}\end{minipage}
 & \begin{minipage}{6.2mm}\centering\textls[-145]{BERT-}\textls[-75]{Score}\end{minipage}
 & \begin{minipage}{6.2mm}\centering\textls[-20]{BLE-URT}\end{minipage}
 & \begin{minipage}{6.2mm}\centering LLa-MA\end{minipage}
 & \begin{minipage}{6.2mm}\centering M-P\end{minipage}
\\ \midrule
\multirow{5}{*}{cs}
 & EuroLLM-9B-Instruct & \human{35.5} & \human{34.3} & \human{31.1} & \human{27.9} & \bleu{2.4} & \chrf{26.2} & \rouge{.186} & \bertscore{.221} & \bleurt{.382} & \llamascore{35.8} & \prometheus{30.4} \\
 & Llama-3.1-8B-Instruct & \human{28.7} & \human{28.3} & \human{27.4} & \human{23.6} & \bleu{3.1} & \chrf{26.5} & \rouge{.212} & \bertscore{.275} & \bleurt{.374} & \llamascore{28.7} & \prometheus{24.0} \\
 & Llama-3.3-70B-Instruct & \human{58.7} & \human{57.4} & \human{56.4} & \human{53.4} & \bleu{4.7} & \chrf{42.4} & \rouge{.365} & \bertscore{.366} & \bleurt{.510} & \llamascore{57.4} & \prometheus{48.9} \\
 & Llama-4-Scout-17B-16E-Instruct & \human{39.2} & \human{37.2} & \human{35.5} & \human{30.6} & \bleu{3.8} & \chrf{32.6} & \rouge{.279} & \bertscore{.323} & \bleurt{.451} & \llamascore{39.1} & \prometheus{30.9} \\
\bottomrule
\end{tabular}

\vspace{10pt}

\begin{tabular}{ll cccc ccccccc}
\toprule
\multicolumn{2}{l}{\multirow{2}{*}{Textual QA: Slovakia}}
& \multicolumn{4}{c}{Manual evaluation}
& \multicolumn{7}{c}{Automatic evaluation}
\\
\cmidrule(lr){3-6}
\cmidrule(lr){7-13}
&
 & \begin{minipage}{6.6mm}\centering Correct \end{minipage}
 & \begin{minipage}{6.6mm}\centering + True    \end{minipage}
 & \begin{minipage}{6.6mm}\centering + \textls[-64]{Relevant}\end{minipage}
 & \begin{minipage}{6.6mm}\centering + \textls[-85]{Coherent}\end{minipage}

 & \begin{minipage}{6.2mm}\centering BL-EU\end{minipage}
 & \begin{minipage}{6.2mm}\centering chrF\end{minipage}
 & \begin{minipage}{6.2mm}\centering\textls[-85]{ROU-GE}\end{minipage}
 & \begin{minipage}{6.2mm}\centering\textls[-145]{BERT-}\textls[-75]{Score}\end{minipage}
 & \begin{minipage}{6.2mm}\centering\textls[-20]{BLE-URT}\end{minipage}
 & \begin{minipage}{6.2mm}\centering LLa-MA\end{minipage}
 & \begin{minipage}{6.2mm}\centering M-P\end{minipage}
\\ \midrule
\multirow{5}{*}{sk}
 & EuroLLM-9B-Instruct & \human{32.0} & \human{30.8} & \human{29.4} & \human{25.4} & \bleu{2.6} & \chrf{24.2} & \rouge{.169} & \bertscore{.677} & \bleurt{.370} & \llamascore{30.6} & \prometheus{27.8} \\
 & Llama-3.1-8B-Instruct & \human{18.9} & \human{18.1} & \human{17.0} & \human{15.0} & \bleu{2.7} & \chrf{21.2} & \rouge{.152} & \bertscore{.689} & \bleurt{.335} & \llamascore{18.5} & \prometheus{13.8} \\
 & Llama-3.3-70B-Instruct & \human{46.2} & \human{45.8} & \human{43.4} & \human{40.8} & \bleu{7.4} & \chrf{39.2} & \rouge{.354} & \bertscore{.755} & \bleurt{.485} & \llamascore{45.4} & \prometheus{38.5} \\
 & Llama-4-Scout-17B-16E-Instruct & \human{34.1} & \human{33.5} & \human{31.8} & \human{30.6} & \bleu{3.9} & \chrf{29.2} & \rouge{.251} & \bertscore{.706} & \bleurt{.425} & \llamascore{32.3} & \prometheus{28.8} \\
\bottomrule
\end{tabular}

\vspace{10pt}

\begin{tabular}{ll cccc ccccccc}
\toprule
\multicolumn{2}{l}{\multirow{2}{*}{Textual QA: Ukraine}}
& \multicolumn{4}{c}{Manual evaluation}
& \multicolumn{7}{c}{Automatic evaluation}
\\
\cmidrule(lr){3-6}
\cmidrule(lr){7-13}
&
 & \begin{minipage}{6.6mm}\centering Correct \end{minipage}
 & \begin{minipage}{6.6mm}\centering + True    \end{minipage}
 & \begin{minipage}{6.6mm}\centering + \textls[-64]{Relevant}\end{minipage}
 & \begin{minipage}{6.6mm}\centering + \textls[-85]{Coherent}\end{minipage}

 & \begin{minipage}{6.2mm}\centering BL-EU\end{minipage}
 & \begin{minipage}{6.2mm}\centering chrF\end{minipage}
 & \begin{minipage}{6.2mm}\centering\textls[-85]{ROU-GE}\end{minipage}
 & \begin{minipage}{6.2mm}\centering\textls[-145]{BERT-}\textls[-75]{Score}\end{minipage}
 & \begin{minipage}{6.2mm}\centering\textls[-20]{BLE-URT}\end{minipage}
 & \begin{minipage}{6.2mm}\centering LLa-MA\end{minipage}
 & \begin{minipage}{6.2mm}\centering M-P\end{minipage}
\\ \midrule
\multirow{5}{*}{uk}
 & EuroLLM-9B-Instruct & \human{26.8} & \human{26.2} & \human{25.7} & \human{22.6} & \bleu{2.4} & \chrf{26.2} & \rouge{.186} & \bertscore{.221} & \bleurt{.382} & \llamascore{35.8} & \prometheus{30.4} \\
 & Llama-3.1-8B-Instruct & \human{22.6} & \human{21.6} & \human{21.3} & \human{20.3} & \bleu{3.1} & \chrf{26.5} & \rouge{.212} & \bertscore{.275} & \bleurt{.374} & \llamascore{28.7} & \prometheus{24.0} \\
 & Llama-3.3-70B-Instruct & \human{51.9} & \human{50.9} & \human{47.3} & \human{46.2} & \bleu{4.7} & \chrf{42.4} & \rouge{.365} & \bertscore{.366} & \bleurt{.510} & \llamascore{57.4} & \prometheus{48.9} \\
 & Llama-4-Scout-17B-16E-Instruct & \human{49.6} & \human{48.1} & \human{46.5} & \human{44.9} & \bleu{3.8} & \chrf{32.6} & \rouge{.279} & \bertscore{.323} & \bleurt{.451} & \llamascore{39.1} & \prometheus{30.9} \\
\bottomrule
\end{tabular}

\caption{Detailed results of textual QA on the development data, including strictness levels of manual evaluation and all tested automatic metrics.}
\label{tab:detailed_baselines_textual}

\end{table}

\begin{table}
\footnotesize\centering
\setlength{\tabcolsep}{4pt}

\begin{tabular}{ll cccc ccccccc}
\toprule
\multicolumn{2}{l}{\multirow{2}{*}{Visual QA: Czechia}}
& \multicolumn{4}{c}{Manual evaluation}
& \multicolumn{7}{c}{Automatic evaluation}
\\
\cmidrule(lr){3-6}
\cmidrule(lr){7-13}
&
 & \begin{minipage}{6.6mm}\centering Correct \end{minipage}
 & \begin{minipage}{6.6mm}\centering + True    \end{minipage}
 & \begin{minipage}{6.6mm}\centering + \textls[-64]{Relevant}\end{minipage}
 & \begin{minipage}{6.6mm}\centering + \textls[-85]{Coherent}\end{minipage}

 & \begin{minipage}{6.2mm}\centering BL-EU\end{minipage}
 & \begin{minipage}{6.2mm}\centering chrF\end{minipage}
 & \begin{minipage}{6.2mm}\centering\textls[-85]{ROU-GE}\end{minipage}
 & \begin{minipage}{6.2mm}\centering\textls[-145]{BERT-}\textls[-75]{Score}\end{minipage}
 & \begin{minipage}{6.2mm}\centering\textls[-20]{BLE-URT}\end{minipage}
 & \begin{minipage}{6.2mm}\centering LLa-MA\end{minipage}
 & \begin{minipage}{6.2mm}\centering M-P\end{minipage}
\\ \midrule
\multirow{5}{*}{cs}
 & gemma3 & \human{12.4} & \human{12.4} & \human{11.5} & \human{8.8} & \bleu{2.7} & \chrf{20.2} & \rouge{.213} & \bertscore{.325} & \bleurt{.359} & \llamascore{11.1} & \prometheus{7.5} \\
 & idefics & \human{5.8} & \human{5.8} & \human{4.4} & \human{3.5} & \bleu{0.9} & \chrf{10.0} & \rouge{.082} & \bertscore{.221} & \bleurt{.209} & \llamascore{7.5} & \prometheus{4.4} \\
 & Llama-3.2-11B-Vision-Instruct & \human{9.7} & \human{9.7} & \human{9.3} & \human{1.8} & \bleu{0.1} & \chrf{9.1} & \rouge{.053} & \bertscore{.093} & \bleurt{.208} & \llamascore{9.7} & \prometheus{8.8} \\
 & Llama-4-Scout-17B-16E-Instruct & \human{14.6} & \human{13.7} & \human{11.9} & \human{11.1} & \bleu{2.7} & \chrf{21.3} & \rouge{.218} & \bertscore{.146} & \bleurt{.356} & \llamascore{14.6} & \prometheus{8.8} \\
 & maya & \human{0.9} & \human{0.9} & \human{0.9} & \human{0.9} & \bleu{0.1} & \chrf{11.6} & \rouge{.082} & \bertscore{.153} & \bleurt{.239} & \llamascore{1.3} & \prometheus{1.8} \\
\bottomrule
\end{tabular}

\vspace{10pt}

\begin{tabular}{ll cccc ccccccc}
\toprule
\multicolumn{2}{l}{\multirow{2}{*}{Visual QA: Slovakia}}
& \multicolumn{4}{c}{Manual evaluation}
& \multicolumn{7}{c}{Automatic evaluation}
\\
\cmidrule(lr){3-6}
\cmidrule(lr){7-13}
&
 & \begin{minipage}{6.6mm}\centering Correct \end{minipage}
 & \begin{minipage}{6.6mm}\centering + True    \end{minipage}
 & \begin{minipage}{6.6mm}\centering + \textls[-64]{Relevant}\end{minipage}
 & \begin{minipage}{6.6mm}\centering + \textls[-85]{Coherent}\end{minipage}

 & \begin{minipage}{6.2mm}\centering BL-EU\end{minipage}
 & \begin{minipage}{6.2mm}\centering chrF\end{minipage}
 & \begin{minipage}{6.2mm}\centering\textls[-85]{ROU-GE}\end{minipage}
 & \begin{minipage}{6.2mm}\centering\textls[-145]{BERT-}\textls[-75]{Score}\end{minipage}
 & \begin{minipage}{6.2mm}\centering\textls[-20]{BLE-URT}\end{minipage}
 & \begin{minipage}{6.2mm}\centering LLa-MA\end{minipage}
 & \begin{minipage}{6.2mm}\centering M-P\end{minipage}
\\ \midrule
\multirow{5}{*}{sk}
 & gemma3 & \human{18.6} & \human{18.6} & \human{18.6} & \human{17.8} & \bleu{9.7} & \chrf{24.9} & \rouge{.202} & \bertscore{.741} & \bleurt{.399} & \llamascore{13.6} & \prometheus{8.5} \\
 & idefics & \human{9.3} & \human{9.3} & \human{8.5} & \human{6.8} & \bleu{3.2} & \chrf{15.7} & \rouge{.116} & \bertscore{.699} & \bleurt{.252} & \llamascore{8.5} & \prometheus{4.2} \\
 & Llama-3.2-11B-Vision-Instruct & \human{12.7} & \human{12.7} & \human{12.7} & \human{0.8} & \bleu{0.6} & \chrf{13.4} & \rouge{.049} & \bertscore{.652} & \bleurt{.200} & \llamascore{10.2} & \prometheus{8.5} \\
 & Llama-4-Scout-17B-16E-Instruct & \human{23.7} & \human{23.7} & \human{23.7} & \human{20.3} & \bleu{4.8} & \chrf{24.4} & \rouge{.214} & \bertscore{.666} & \bleurt{.412} & \llamascore{18.6} & \prometheus{10.2} \\
 & maya & \human{0.8} & \human{0.8} & \human{0.8} & \human{0.8} & \bleu{0.5} & \chrf{15.6} & \rouge{.087} & \bertscore{.667} & \bleurt{.240} & \llamascore{0.0} & \prometheus{0.0} \\
\bottomrule
\end{tabular}

\vspace{10pt}

\begin{tabular}{ll cccc ccccccc}
\toprule
\multicolumn{2}{l}{\multirow{2}{*}{Visual QA: Ukraine}}
& \multicolumn{4}{c}{Manual evaluation}
& \multicolumn{7}{c}{Automatic evaluation}
\\
\cmidrule(lr){3-6}
\cmidrule(lr){7-13}
&
 & \begin{minipage}{6.6mm}\centering Correct \end{minipage}
 & \begin{minipage}{6.6mm}\centering + True    \end{minipage}
 & \begin{minipage}{6.6mm}\centering + \textls[-64]{Relevant}\end{minipage}
 & \begin{minipage}{6.6mm}\centering + \textls[-85]{Coherent}\end{minipage}

 & \begin{minipage}{6.2mm}\centering BL-EU\end{minipage}
 & \begin{minipage}{6.2mm}\centering chrF\end{minipage}
 & \begin{minipage}{6.2mm}\centering\textls[-85]{ROU-GE}\end{minipage}
 & \begin{minipage}{6.2mm}\centering\textls[-145]{BERT-}\textls[-75]{Score}\end{minipage}
 & \begin{minipage}{6.2mm}\centering\textls[-20]{BLE-URT}\end{minipage}
 & \begin{minipage}{6.2mm}\centering LLa-MA\end{minipage}
 & \begin{minipage}{6.2mm}\centering M-P\end{minipage}
\\ \midrule
\multirow{5}{*}{uk}
 & gemma3 & \human{25.5} & \human{25.5} & \human{25.5} & \human{24.0} & \bleu{4.2} & \chrf{19.9} & \rouge{.026} & \bertscore{.721} & \bleurt{.338} & \llamascore{21.6} & \prometheus{14.7} \\
 & idefics & \human{7.8} & \human{7.8} & \human{7.8} & \human{7.4} & \bleu{0.8} & \chrf{4.4} & \rouge{.005} & \bertscore{.666} & \bleurt{.156} & \llamascore{7.8} & \prometheus{3.9} \\
 & Llama-3.2-11B-Vision-Instruct & \human{16.2} & \human{16.2} & \human{16.2} & \human{8.3} & \bleu{0.3} & \chrf{6.5} & \rouge{.005} & \bertscore{.647} & \bleurt{.197} & \llamascore{15.2} & \prometheus{10.3} \\
 & Llama-4-Scout-17B-16E-Instruct & \human{29.9} & \human{29.9} & \human{29.9} & \human{29.9} & \bleu{2.1} & \chrf{21.1} & \rouge{.023} & \bertscore{.663} & \bleurt{.357} & \llamascore{26.0} & \prometheus{15.2} \\
 & maya & \human{3.4} & \human{3.4} & \human{3.4} & \human{2.5} & \bleu{0.2} & \chrf{12.3} & \rouge{.010} & \bertscore{.656} & \bleurt{.171} & \llamascore{2.9} & \prometheus{2.0} \\
\bottomrule
\end{tabular}

\caption{Detailed results on the visual QA on development data, including strictness levels of manual evaluation and all tested automatic metrics.}\label{tab:detailed_baselines_visual}

\end{table}



\begin{table}
\footnotesize
\setlength{\tabcolsep}{1.3pt}

\begin{tabular}{ll cccc @{\hspace{4pt}} cccc @{\hspace{4pt}} cccc @{\hspace{4pt}} cccc @{\hspace{4pt}} cccc @{\hspace{4pt}} cccc @{\hspace{4pt}} }
\toprule
& \multirow{2}{*}{Model}  & \multicolumn{4}{c}{Czechia (cs)} & \multicolumn{4}{c}{Czechia (en)} & \multicolumn{4}{c}{Slovakia (sk)} & \multicolumn{4}{c}{Slovakia (en)} & \multicolumn{4}{c}{Ukraine (uk)} & \multicolumn{4}{c}{Ukraine (en)} \\
\cmidrule(lr){3-10} \cmidrule(lr){11-18} \cmidrule(lr){19-26} 
 &  & geo  & cul  & his  & oth  & geo  & cul  & his  & oth  & geo  & cul  & his  & oth  & geo  & cul  & his  & oth  & geo  & cul  & his  & oth  & geo  & cul  & his  & oth  \\
\midrule
\multirow{5}{*}{\rotatebox{90}{Textual}}~ & EuroLLM 9B Ins. & \human{40} & \human{14} & \human{33} & \human{12} & \human{33} & \human{12} & \human{29} & \human{12} & \human{30} & \human{16} & \human{37} & \human{20} & \human{30} & \human{15} & \human{35} & \human{22} & \human{30} & \human{11} & \human{24} & \human{21} & \human{23} & \human{14} & \human{26} & \human{19} \\
& LLaMA 3.1 8B Ins. & \human{32} & \human{9} & \human{31} & \human{15} & \human{34} & \human{13} & \human{37} & \human{14} & \human{15} & \human{8} & \human{15} & \human{23} & \human{27} & \human{10} & \human{21} & \human{27} & \human{25} & \human{8} & \human{23} & \human{21} & \human{31} & \human{19} & \human{25} & \human{31} \\
& LLaMA 3.3 70B Ins. & \human{65} & \human{41} & \human{57} & \human{38} & \human{58} & \human{34} & \human{44} & \human{28} & \human{49} & \human{29} & \human{44} & \human{35} & \human{49} & \human{24} & \human{44} & \human{38} & \human{52} & \human{41} & \human{42} & \human{46} & \human{49} & \human{34} & \human{39} & \human{46} \\
\bottomrule
\end{tabular}

\caption{Per-category breakdown how human evaluation scores (on the strictest level) fro evaluted model. We considered geography, culture and history. Politics, sports and other were put in the same category because they would be too sparse.}\label{tab:appendix_breakdown}

\end{table}


\begin{table}
\footnotesize\centering
\setlength{\tabcolsep}{4pt}

\begin{tabular}{ll c c c c c c c c c c c c c c c c c c c c}
\toprule
\multirow{2}{*}{\rotatebox{90}{Loc.}} & \multirow{2}{*}{Model} & \multicolumn{10}{c}{Local language} & \multicolumn{10}{c}{English} \\

    \cmidrule(lr){3-12} \cmidrule(lr){13-22}

    & & \multicolumn{5}{c}{content} & \multicolumn{2}{c}{text} & \multicolumn{3}{c}{medium} & \multicolumn{5}{c}{content} & \multicolumn{2}{c}{text} & \multicolumn{3}{c}{medium} \\
    \cmidrule(lr){3-7} \cmidrule(lr){8-9} \cmidrule(lr){10-12} \cmidrule(lr){13-17} \cmidrule(lr){18-19} \cmidrule(lr){20-22}
 &  & \rotatebox{90}{portrait} & \rotatebox{90}{building} & \rotatebox{90}{city} & \rotatebox{90}{nature} & \rotatebox{90}{other} & \rotatebox{90}{text} & \rotatebox{90}{no text} & \rotatebox{90}{color photo} & \rotatebox{90}{BW photo} & \rotatebox{90}{painting} & \rotatebox{90}{portrait} & \rotatebox{90}{building} & \rotatebox{90}{city} & \rotatebox{90}{nature} & \rotatebox{90}{other} & \rotatebox{90}{text} & \rotatebox{90}{no text} & \rotatebox{90}{color photo} & \rotatebox{90}{BW photo} & \rotatebox{90}{painting} \\
\midrule
\multirow{5}{*}{\rotatebox{90}{Czechia}} & gemma3 & \human{8} & \human{6} & \human{9} & \human{0} & \human{28} & \human{15} & \human{8} & \human{9} & \human{9} & \human{12} & \human{0} & \human{5} & \human{3} & \human{0} & \human{16} & \human{8} & \human{4} & \human{5} & \human{0} & \human{12} \\
& idefics & \human{2} & \human{2} & \human{4} & \human{6} & \human{12} & \human{15} & \human{1} & \human{3} & \human{4} & \human{12} & \human{2} & \human{1} & \human{1} & \human{0} & \human{4} & \human{5} & \human{1} & \human{1} & \human{4} & \human{0} \\
& LLaMA 3.2 11B Vis. Ins. & \human{0} & \human{2} & \human{1} & \human{0} & \human{8} & \human{5} & \human{1} & \human{2} & \human{0} & \human{0} & \human{0} & \human{8} & \human{12} & \human{12} & \human{16} & \human{10} & \human{8} & \human{9} & \human{4} & \human{0} \\
\bottomrule
\end{tabular}

\caption{Breakdown of the model accuracy in human evaluation based on the visual content of the images. Categories of food and technology are merged into others due to a low number of examples.}\label{tab:appendix_visual_breakdown}

\end{table}


\begin{table}

\footnotesize\centering
\setlength{\tabcolsep}{3pt}
\newcolumntype{E}{>{\centering\arraybackslash}p{8mm}}
\begin{tabular}{l *{14}{E}}
\toprule
 & \multicolumn{7}{c}{Textual} & \multicolumn{7}{c}{Visual} \\
\cmidrule(lr){2-8} \cmidrule(lr){9-15}
 & BLEU & chrF & RG & BScr & BLRT & LLM & M-P & BLEU & chrF & RG & BScr & BLRT & LLM & M-P \\
 \midrule
Correct    & \Co{.884} & \Co{.927} & \Co{.917} & \Co{.857} & \Co{.931} & \Co{.963} & \Co{.951} & \Co{.577} & \Co{.670} & \Co{.842} & \Co{-.108} & \Co{.892} & \Co{.990} & \Co{.961} \\
+ true     & \Co{.897} & \Co{.943} & \Co{.933} & \Co{.877} & \Co{.937} & \Co{.964} & \Co{.955} & \Co{.577} & \Co{.668} & \Co{.839} & \Co{-.098} & \Co{.890} & \Co{.989} & \Co{.962} \\
+ relevant & \Co{.904} & \Co{.954} & \Co{.941} & \Co{.897} & \Co{.920} & \Co{.945} & \Co{.943} & \Co{.571} & \Co{.668} & \Co{.835} & \Co{-.114} & \Co{.888} & \Co{.981} & \Co{.965} \\
+ coherent & \Co{.907} & \Co{.956} & \Co{.944} & \Co{.905} & \Co{.923} & \Co{.942} & \Co{.940} & \Co{.605} & \Co{.773} & \Co{.882} & \Co{-.039} & \Co{.932} & \Co{.908} & \Co{.848} \\
\bottomrule
\end{tabular}

\caption{System-level Pearson correlations of the automatic evaluation metrics with different levels of strictness of human evaluation. The presented numbers are averages of Pearson correlations computed separately for each country and language.}\label{tab:appendix_pearson}

\end{table}

\begin{table}

Textual QA

\resizebox{.33\textwidth}{!}{%
\footnotesize\centering
\setlength{\tabcolsep}{4pt}
\begin{tabular}{l cccccccc}
\toprule
\Large CZ: cs
 & \begin{minipage}{6.0mm}\centering BL-EU\end{minipage}
 & \begin{minipage}{6.0mm}\centering\ chrF\end{minipage}
 & \begin{minipage}{6.0mm}\centering\textls[-85]{ROU-GE}\end{minipage}
 & \begin{minipage}{6.0mm}\centering\textls[-145]{BERT-}\textls[-75]{Score}\end{minipage}
 & \begin{minipage}{6.0mm}\centering\textls[-20]{BLE-URT}\end{minipage}
 & \begin{minipage}{6.0mm}\centering LLa-MA\end{minipage}
 & \begin{minipage}{6.0mm}\centering M-P\end{minipage}
 \\
 \midrule
Correct    & \Co{.90} & \Co{.90} & \Co{.90} & \Co{.90} & \Co{1.00} & \Co{1.00} & \Co{1.00} \\
+ true     & \Co{.90} & \Co{.90} & \Co{.90} & \Co{.90} & \Co{1.00} & \Co{1.00} & \Co{1.00} \\
+ relevant & \Co{.90} & \Co{.90} & \Co{.90} & \Co{.90} & \Co{1.00} & \Co{1.00} & \Co{1.00} \\
+ coherent & \Co{.90} & \Co{.90} & \Co{.90} & \Co{.90} & \Co{1.00} & \Co{1.00} & \Co{1.00} \\

\bottomrule
\end{tabular}}
\resizebox{.33\textwidth}{!}{%
\footnotesize\centering
\setlength{\tabcolsep}{4pt}
\begin{tabular}{l cccccccc}
\toprule
\Large SK: sk
 & \begin{minipage}{6.0mm}\centering BL-EU\end{minipage}
 & \begin{minipage}{6.0mm}\centering\ chrF\end{minipage}
 & \begin{minipage}{6.0mm}\centering\textls[-85]{ROU-GE}\end{minipage}
 & \begin{minipage}{6.0mm}\centering\textls[-145]{BERT-}\textls[-75]{Score}\end{minipage}
 & \begin{minipage}{6.0mm}\centering\textls[-20]{BLE-URT}\end{minipage}
 & \begin{minipage}{6.0mm}\centering LLa-MA\end{minipage}
 & \begin{minipage}{6.0mm}\centering M-P\end{minipage}
 \\
 \midrule
Correct    & \Co{.90} & \Co{1.00} & \Co{1.00} & \Co{.90} & \Co{1.00} & \Co{1.00} & \Co{1.00} \\
+ true     & \Co{.90} & \Co{1.00} & \Co{1.00} & \Co{.90} & \Co{1.00} & \Co{1.00} & \Co{1.00} \\
+ relevant & \Co{.90} & \Co{1.00} & \Co{1.00} & \Co{.90} & \Co{1.00} & \Co{1.00} & \Co{1.00} \\
+ coherent & \Co{.90} & \Co{1.00} & \Co{1.00} & \Co{.90} & \Co{1.00} & \Co{1.00} & \Co{1.00} \\
\bottomrule
\end{tabular}}
\resizebox{.33\textwidth}{!}{%
\footnotesize\centering
\setlength{\tabcolsep}{4pt}
\begin{tabular}{l cccccccc}
\toprule
\Large UA: uk
 & \begin{minipage}{6.0mm}\centering BL-EU\end{minipage}
 & \begin{minipage}{6.0mm}\centering\ chrF\end{minipage}
 & \begin{minipage}{6.0mm}\centering\textls[-85]{ROU-GE}\end{minipage}
 & \begin{minipage}{6.0mm}\centering\textls[-145]{BERT-}\textls[-75]{Score}\end{minipage}
 & \begin{minipage}{6.0mm}\centering\textls[-20]{BLE-URT}\end{minipage}
 & \begin{minipage}{6.0mm}\centering LLa-MA\end{minipage}
 & \begin{minipage}{6.0mm}\centering M-P\end{minipage}
 \\
 \midrule

Correct    & \Co{.90} & \Co{.90} & \Co{.90} & \Co{.90} & \Co{1.00} & \Co{1.00} & \Co{1.00} \\
+ true     & \Co{.90} & \Co{.90} & \Co{.90} & \Co{.90} & \Co{1.00} & \Co{1.00} & \Co{1.00} \\
+ relevant & \Co{.90} & \Co{.90} & \Co{.90} & \Co{.90} & \Co{1.00} & \Co{1.00} & \Co{1.00} \\
+ coherent & \Co{.90} & \Co{.90} & \Co{.90} & \Co{.90} & \Co{1.00} & \Co{1.00} & \Co{1.00} \\

\bottomrule
\end{tabular}}

\vspace{5pt}

\resizebox{.33\textwidth}{!}{%
\footnotesize\centering
\setlength{\tabcolsep}{4pt}
\begin{tabular}{l cccccccc}
\toprule
\Large CZ: en
 & \begin{minipage}{6.0mm}\centering BL-EU\end{minipage}
 & \begin{minipage}{6.0mm}\centering\ chrF\end{minipage}
 & \begin{minipage}{6.0mm}\centering\textls[-85]{ROU-GE}\end{minipage}
 & \begin{minipage}{6.0mm}\centering\textls[-145]{BERT-}\textls[-75]{Score}\end{minipage}
 & \begin{minipage}{6.0mm}\centering\textls[-20]{BLE-URT}\end{minipage}
 & \begin{minipage}{6.0mm}\centering LLa-MA\end{minipage}
 & \begin{minipage}{6.0mm}\centering M-P\end{minipage}
 \\
 \midrule
Correct    & \Co{.60} & \Co{.70} & \Co{.70} & \Co{.70} & \Co{.50} & \Co{1.00} & \Co{1.00} \\
+ true     & \Co{.60} & \Co{.70} & \Co{.70} & \Co{.70} & \Co{.50} & \Co{1.00} & \Co{1.00} \\
+ relevant & \Co{.70} & \Co{.90} & \Co{.90} & \Co{.90} & \Co{.80} & \Co{.90 }&  \Co{ .90} \\
+ coherent & \Co{.70} & \Co{.90} & \Co{.90} & \Co{.90} & \Co{.80} & \Co{.90 }&  \Co{ .90}  \\
\bottomrule
\end{tabular}}
\resizebox{.33\textwidth}{!}{%
\footnotesize\centering
\setlength{\tabcolsep}{4pt}
\begin{tabular}{l cccccccc}
\toprule
\Large SK: en
 & \begin{minipage}{6.0mm}\centering BL-EU\end{minipage}
 & \begin{minipage}{6.0mm}\centering\ chrF\end{minipage}
 & \begin{minipage}{6.0mm}\centering\textls[-85]{ROU-GE}\end{minipage}
 & \begin{minipage}{6.0mm}\centering\textls[-145]{BERT-}\textls[-75]{Score}\end{minipage}
 & \begin{minipage}{6.0mm}\centering\textls[-20]{BLE-URT}\end{minipage}
 & \begin{minipage}{6.0mm}\centering LLa-MA\end{minipage}
 & \begin{minipage}{6.0mm}\centering M-P\end{minipage}
 \\
 \midrule
 Correct   & \Co{.60} & \Co{.90} & \Co{.70} & \Co{.70} & \Co{.90} & \Co{1.00} & \Co{1.00} \\
+ true     & \Co{.60} & \Co{.90} & \Co{.70} & \Co{.70} & \Co{.90} & \Co{1.00} & \Co{1.00} \\
+ relevant & \Co{.60} & \Co{.90} & \Co{.70} & \Co{.70} & \Co{.90} & \Co{1.00} & \Co{1.00} \\
+ coherent & \Co{.60} & \Co{.90} & \Co{.70} & \Co{.70} & \Co{.90} & \Co{1.00} & \Co{1.00} \\
\bottomrule
\end{tabular}}
\resizebox{.33\textwidth}{!}{%
\footnotesize\centering
\setlength{\tabcolsep}{4pt}
\begin{tabular}{l cccccccc}
\toprule
\Large UA: en
 & \begin{minipage}{6.0mm}\centering BL-EU\end{minipage}
 & \begin{minipage}{6.0mm}\centering\ chrF\end{minipage}
 & \begin{minipage}{6.0mm}\centering\textls[-85]{ROU-GE}\end{minipage}
 & \begin{minipage}{6.0mm}\centering\textls[-145]{BERT-}\textls[-75]{Score}\end{minipage}
 & \begin{minipage}{6.0mm}\centering\textls[-20]{BLE-URT}\end{minipage}
 & \begin{minipage}{6.0mm}\centering LLa-MA\end{minipage}
 & \begin{minipage}{6.0mm}\centering M-P\end{minipage}
 \\
 \midrule
Correct    & \Co{.50} & \Co{.80} & \Co{.80} & \Co{.80} & \Co{.90} & \Co{.70} & \Co{.70} \\
+ true     & \Co{.62} & \Co{.87} & \Co{.87} & \Co{.87} & \Co{.87} & \Co{.82} & \Co{.82} \\
+ relevant & \Co{.70} & \Co{.90} & \Co{.90} & \Co{.90} & \Co{.80} & \Co{.90} & \Co{.90} \\
+ coherent & \Co{.70} & \Co{.90} & \Co{.90} & \Co{.90} & \Co{.80} & \Co{.90} & \Co{.90}  \\
\bottomrule
\end{tabular}}

\vspace{15pt}
Visual QA

\resizebox{.33\textwidth}{!}{%
\footnotesize\centering
\setlength{\tabcolsep}{4pt}
\begin{tabular}{l cccccccc}
\toprule
\Large CZ: cs
 & \begin{minipage}{6.0mm}\centering BL-EU\end{minipage}
 & \begin{minipage}{6.0mm}\centering\ chrF\end{minipage}
 & \begin{minipage}{6.0mm}\centering\textls[-85]{ROU-GE}\end{minipage}
 & \begin{minipage}{6.0mm}\centering\textls[-145]{BERT-}\textls[-75]{Score}\end{minipage}
 & \begin{minipage}{6.0mm}\centering\textls[-20]{BLE-URT}\end{minipage}
 & \begin{minipage}{6.0mm}\centering LLa-MA\end{minipage}
 & \begin{minipage}{6.0mm}\centering M-P\end{minipage}
 \\
 \midrule

 Correct   & \Co{ .90} & \Co{.60} & \Co{.60} & \Co{-.10} & \Co{.50} & \Co{1.00} & \Co{.82} \\
+ true     & \Co{ .90} & \Co{.60} & \Co{.60} & \Co{-.10} & \Co{.50} & \Co{1.00} & \Co{.82} \\
+ relevant & \Co{ .90} & \Co{.60} & \Co{.60} & \Co{-.10} & \Co{.50} & \Co{1.00} & \Co{.82} \\
+ coherent & \Co{1.00} & \Co{.70} & \Co{.70} & \Co{ .20} & \Co{.60} & \Co{ .90} & \Co{.56} \\

\bottomrule
\end{tabular}}
\resizebox{.33\textwidth}{!}{%
\footnotesize\centering
\setlength{\tabcolsep}{4pt}
\begin{tabular}{l cccccccc}
\toprule
\Large SK: sk
 & \begin{minipage}{6.0mm}\centering BL-EU\end{minipage}
 & \begin{minipage}{6.0mm}\centering\ chrF\end{minipage}
 & \begin{minipage}{6.0mm}\centering\textls[-85]{ROU-GE}\end{minipage}
 & \begin{minipage}{6.0mm}\centering\textls[-145]{BERT-}\textls[-75]{Score}\end{minipage}
 & \begin{minipage}{6.0mm}\centering\textls[-20]{BLE-URT}\end{minipage}
 & \begin{minipage}{6.0mm}\centering LLa-MA\end{minipage}
 & \begin{minipage}{6.0mm}\centering M-P\end{minipage}
 \\
 \midrule

Correct    & \Co{.80} & \Co{.60} & \Co{.70} & \Co{-.10} & \Co{.70} & \Co{1.00} & \Co{.97} \\
+ true     & \Co{.80} & \Co{.60} & \Co{.70} & \Co{-.10} & \Co{.70} & \Co{1.00} & \Co{.97} \\
+ relevant & \Co{.80} & \Co{.60} & \Co{.70} & \Co{-.10} & \Co{.70} & \Co{1.00} & \Co{.97} \\
+ coherent & \Co{.87} & \Co{.87} & \Co{.97} & \Co{ .31} & \Co{.97} & \Co{ .82} & \Co{.71} \\

\bottomrule
\end{tabular}}
\resizebox{.33\textwidth}{!}{%
\footnotesize\centering
\setlength{\tabcolsep}{4pt}
\begin{tabular}{l cccccccc}
\toprule
\Large UA: uk
 & \begin{minipage}{6.0mm}\centering BL-EU\end{minipage}
 & \begin{minipage}{6.0mm}\centering\ chrF\end{minipage}
 & \begin{minipage}{6.0mm}\centering\textls[-85]{ROU-GE}\end{minipage}
 & \begin{minipage}{6.0mm}\centering\textls[-145]{BERT-}\textls[-75]{Score}\end{minipage}
 & \begin{minipage}{6.0mm}\centering\textls[-20]{BLE-URT}\end{minipage}
 & \begin{minipage}{6.0mm}\centering LLa-MA\end{minipage}
 & \begin{minipage}{6.0mm}\centering M-P\end{minipage}
 \\
 \midrule

Correct    & \Co{.80} & \Co{.70} & \Co{.56} & \Co{.30} & \Co{.90} & \Co{1.00} & \Co{1.00} \\
+ true     & \Co{.80} & \Co{.70} & \Co{.56} & \Co{.30} & \Co{.90} & \Co{1.00} & \Co{1.00} \\
+ relevant & \Co{.80} & \Co{.70} & \Co{.56} & \Co{.30} & \Co{.90} & \Co{1.00} & \Co{1.00} \\
+ coherent & \Co{.80} & \Co{.70} & \Co{.56} & \Co{.30} & \Co{.90} & \Co{1.00} & \Co{1.00} \\

\bottomrule
\end{tabular}}

\vspace{5pt}

\resizebox{.33\textwidth}{!}{%
\footnotesize\centering
\setlength{\tabcolsep}{4pt}
\begin{tabular}{l cccccccc}
\toprule
\Large CZ: en
 & \begin{minipage}{6.0mm}\centering BL-EU\end{minipage}
 & \begin{minipage}{6.0mm}\centering\ chrF\end{minipage}
 & \begin{minipage}{6.0mm}\centering\textls[-85]{ROU-GE}\end{minipage}
 & \begin{minipage}{6.0mm}\centering\textls[-145]{BERT-}\textls[-75]{Score}\end{minipage}
 & \begin{minipage}{6.0mm}\centering\textls[-20]{BLE-URT}\end{minipage}
 & \begin{minipage}{6.0mm}\centering LLa-MA\end{minipage}
 & \begin{minipage}{6.0mm}\centering M-P\end{minipage}
 \\
 \midrule

Correct & \Co{.70} & \Co{.60} & \Co{.80} & \Co{-.70} & \Co{.90} & \Co{1.00} & \Co{1.00} \\
+ true & \Co{.70} & \Co{.60} & \Co{.80} & \Co{-.70} & \Co{.90} & \Co{1.00} & \Co{1.00} \\
+ relevant & \Co{.70} & \Co{.60} & \Co{.80} & \Co{-.70} & \Co{.90} & \Co{1.00} & \Co{1.00} \\
+ coherent & \Co{.70} & \Co{.60} & \Co{.80} & \Co{-.70} & \Co{.90} & \Co{1.00} & \Co{1.00} \\

\bottomrule
\end{tabular}}
\resizebox{.33\textwidth}{!}{%
\footnotesize\centering
\setlength{\tabcolsep}{4pt}
\begin{tabular}{l cccccccc}
\toprule
\Large SK: en
 & \begin{minipage}{6.0mm}\centering BL-EU\end{minipage}
 & \begin{minipage}{6.0mm}\centering\ chrF\end{minipage}
 & \begin{minipage}{6.0mm}\centering\textls[-85]{ROU-GE}\end{minipage}
 & \begin{minipage}{6.0mm}\centering\textls[-145]{BERT-}\textls[-75]{Score}\end{minipage}
 & \begin{minipage}{6.0mm}\centering\textls[-20]{BLE-URT}\end{minipage}
 & \begin{minipage}{6.0mm}\centering LLa-MA\end{minipage}
 & \begin{minipage}{6.0mm}\centering M-P\end{minipage}
 \\
 \midrule

Correct & \Co{.20} & \Co{.30} & \Co{.90} & \Co{-.50} & \Co{.90} & \Co{1.00} & \Co{.97} \\
+ true & \Co{.20} & \Co{.30} & \Co{.90} & \Co{-.50} & \Co{.90} & \Co{1.00} & \Co{.97} \\
+ relevant & \Co{.20} & \Co{.30} & \Co{.90} & \Co{-.50} & \Co{.90} & \Co{1.00} & \Co{.97} \\
+ coherent & \Co{.10} & \Co{.60} & \Co{.70} & \Co{-.70} & \Co{.70} & \Co{.90} & \Co{.97} \\

\bottomrule
\end{tabular}}
\resizebox{.33\textwidth}{!}{%
\footnotesize\centering
\setlength{\tabcolsep}{4pt}
\begin{tabular}{l cccccccc}
\toprule
\Large UA: en
 & \begin{minipage}{6.0mm}\centering BL-EU\end{minipage}
 & \begin{minipage}{6.0mm}\centering\ chrF\end{minipage}
 & \begin{minipage}{6.0mm}\centering\textls[-85]{ROU-GE}\end{minipage}
 & \begin{minipage}{6.0mm}\centering\textls[-145]{BERT-}\textls[-75]{Score}\end{minipage}
 & \begin{minipage}{6.0mm}\centering\textls[-20]{BLE-URT}\end{minipage}
 & \begin{minipage}{6.0mm}\centering LLa-MA\end{minipage}
 & \begin{minipage}{6.0mm}\centering M-P\end{minipage}
 \\
 \midrule

Correct & \Co{.60} & \Co{.50} & \Co{1.00} & \Co{-.50} & \Co{1.00} & \Co{1.00} & \Co{1.00} \\
+ true & \Co{.60} & \Co{.50} & \Co{1.00} & \Co{-.50} & \Co{1.00} & \Co{1.00} & \Co{1.00} \\
+ relevant & \Co{.60} & \Co{.50} & \Co{1.00} & \Co{-.50} & \Co{1.00} & \Co{1.00} & \Co{1.00} \\
+ coherent & \Co{.30} & \Co{.70} & \Co{.90} & \Co{-.70} & \Co{.90} & \Co{.90} & \Co{.90} \\

\bottomrule
\end{tabular}}

\caption{System-level Spearman correlation of automatic metrics and human judgment split over countries and languages. Note that each of the tables is only based on a comparison of 5 systems.}\label{tab:system_level_individual}

\end{table}

\begin{table}

Textual QA \\
\resizebox{.33\textwidth}{!}{%
\footnotesize\centering
\setlength{\tabcolsep}{4pt}
\begin{tabular}{l cccccccc}
\toprule
\Large CZ: cs
 & \begin{minipage}{9.0mm}\centering chrF\end{minipage}
 & \begin{minipage}{9.0mm}\centering ROU-GE\end{minipage}
 & \begin{minipage}{9.0mm}\centering BLE-URT\end{minipage}
 & \begin{minipage}{9.0mm}\centering LLa-MA\end{minipage}
 & \begin{minipage}{9.0mm}\centering M-P\end{minipage}
 \\
 \midrule

Correct & \R{.716} & \R{.656} & \R{.682} & \R{.893} & \R{.795} \\
+ true & \R{.707} & \R{.649} & \R{.671} & \R{.879} & \R{.793} \\
+ relevant & \R{.705} & \R{.647} & \R{.658} & \R{.859} & \R{.793} \\
+ coherent & \R{.664} & \R{.629} & \R{.622} & \R{.801} & \R{.756} \\

\bottomrule
\end{tabular}}
\resizebox{.33\textwidth}{!}{%
\footnotesize\centering
\setlength{\tabcolsep}{4pt}
\begin{tabular}{l cccccccc}
\toprule
\Large SK: sk
 & \begin{minipage}{9.0mm}\centering chrF\end{minipage}
 & \begin{minipage}{9.0mm}\centering ROU-GE\end{minipage}
 & \begin{minipage}{9.0mm}\centering BLE-URT\end{minipage}
 & \begin{minipage}{9.0mm}\centering LLa-MA\end{minipage}
 & \begin{minipage}{9.0mm}\centering M-P\end{minipage}
 \\
 \midrule

Correct & \R{.649} & \R{.575} & \R{.562} & \R{.867} & \R{.827} \\
+ true & \R{.645} & \R{.578} & \R{.564} & \R{.854} & \R{.820} \\
+ relevant & \R{.621} & \R{.554} & \R{.542} & \R{.832} & \R{.803} \\
+ coherent & \R{.591} & \R{.543} & \R{.530} & \R{.797} & \R{.789} \\

\bottomrule
\end{tabular}}
\resizebox{.33\textwidth}{!}{%
\footnotesize\centering
\setlength{\tabcolsep}{4pt}
\begin{tabular}{l cccccccc}
\toprule
\Large UA: uk
 & \begin{minipage}{9.0mm}\centering chrF\end{minipage}
 & \begin{minipage}{9.0mm}\centering ROU-GE\end{minipage}
 & \begin{minipage}{9.0mm}\centering BLE-URT\end{minipage}
 & \begin{minipage}{9.0mm}\centering LLa-MA\end{minipage}
 & \begin{minipage}{9.0mm}\centering M-P\end{minipage}
 \\
 \midrule

Correct & \R{.625} & \R{.147} & \R{.577} & \R{.884} & \R{.788} \\
+ true & \R{.617} & \R{.145} & \R{.565} & \R{.877} & \R{.791} \\
+ relevant & \R{.616} & \R{.144} & \R{.557} & \R{.856} & \R{.780} \\
+ coherent & \R{.601} & \R{.138} & \R{.544} & \R{.815} & \R{.751} \\

\bottomrule
\end{tabular}}

\resizebox{.33\textwidth}{!}{%
\footnotesize\centering
\setlength{\tabcolsep}{4pt}
\begin{tabular}{l cccccccc}
\toprule
\Large CZ: en
 & \begin{minipage}{9.0mm}\centering chrF\end{minipage}
 & \begin{minipage}{9.0mm}\centering ROU-GE\end{minipage}
 & \begin{minipage}{9.0mm}\centering BLE-URT\end{minipage}
 & \begin{minipage}{9.0mm}\centering LLa-MA\end{minipage}
 & \begin{minipage}{9.0mm}\centering M-P\end{minipage}
 \\
 \midrule

Correct & \R{.571} & \R{.534} & \R{.554} & \R{.714} & \R{.641} \\
+ true & \R{.558} & \R{.528} & \R{.535} & \R{.690} & \R{.639} \\
+ relevant & \R{.527} & \R{.506} & \R{.506} & \R{.646} & \R{.608} \\
+ coherent & \R{.514} & \R{.495} & \R{.494} & \R{.628} & \R{.594} \\

\bottomrule
\end{tabular}}
\resizebox{.33\textwidth}{!}{%
\footnotesize\centering
\setlength{\tabcolsep}{4pt}
\begin{tabular}{l cccccccc}
\toprule
\Large SK: en
 & \begin{minipage}{9.0mm}\centering chrF\end{minipage}
 & \begin{minipage}{9.0mm}\centering ROU-GE\end{minipage}
 & \begin{minipage}{9.0mm}\centering BLE-URT\end{minipage}
 & \begin{minipage}{9.0mm}\centering LLa-MA\end{minipage}
 & \begin{minipage}{9.0mm}\centering M-P\end{minipage}
 \\
 \midrule

Correct & \R{.515} & \R{.520} & \R{.489} & \R{.712} & \R{.637} \\
+ true & \R{.509} & \R{.517} & \R{.480} & \R{.695} & \R{.625} \\
+ relevant & \R{.475} & \R{.489} & \R{.461} & \R{.659} & \R{.600} \\
+ coherent & \R{.465} & \R{.482} & \R{.454} & \R{.647} & \R{.595} \\

\bottomrule
\end{tabular}}
\resizebox{.33\textwidth}{!}{%
\footnotesize\centering
\setlength{\tabcolsep}{4pt}
\begin{tabular}{l cccccccc}
\toprule
\Large UA: en
 & \begin{minipage}{9.0mm}\centering chrF\end{minipage}
 & \begin{minipage}{9.0mm}\centering ROU-GE\end{minipage}
 & \begin{minipage}{9.0mm}\centering BLE-URT\end{minipage}
 & \begin{minipage}{9.0mm}\centering LLa-MA\end{minipage}
 & \begin{minipage}{9.0mm}\centering M-P\end{minipage}
 \\
 \midrule

Correct & \R{.485} & \R{.430} & \R{.450} & \R{.671} & \R{.556} \\
+ true & \R{.475} & \R{.417} & \R{.442} & \R{.652} & \R{.552} \\
+ relevant & \R{.460} & \R{.401} & \R{.422} & \R{.606} & \R{.512} \\
+ coherent & \R{.454} & \R{.400} & \R{.419} & \R{.599} & \R{.506} \\

\bottomrule
\end{tabular}}

\vspace{10pt}

Visual QA \\
\resizebox{.33\textwidth}{!}{%
\footnotesize\centering
\setlength{\tabcolsep}{4pt}
\begin{tabular}{l cccccccc}
\toprule
\Large CZ: cs
 & \begin{minipage}{9.0mm}\centering chrF\end{minipage}
 & \begin{minipage}{9.0mm}\centering ROU-GE\end{minipage}
 & \begin{minipage}{9.0mm}\centering BLE-URT\end{minipage}
 & \begin{minipage}{9.0mm}\centering LLa-MA\end{minipage}
 & \begin{minipage}{9.0mm}\centering M-P\end{minipage}
 \\
 \midrule

 Correct & \R{.374} & \R{.334} & \R{.363} & \R{.807} & \R{.638} \\
+ true & \R{.376} & \R{.335} & \R{.359} & \R{.795} & \R{.631} \\
+ relevant & \R{.404} & \R{.359} & \R{.390} & \R{.799} & \R{.672} \\
+ coherent & \R{.365} & \R{.378} & \R{.349} & \R{.679} & \R{.602} \\

\bottomrule
\end{tabular}}
\resizebox{.33\textwidth}{!}{%
\footnotesize\centering
\setlength{\tabcolsep}{4pt}
\begin{tabular}{l cccccccc}
\toprule
\Large SK: sk
 & \begin{minipage}{9.0mm}\centering chrF\end{minipage}
 & \begin{minipage}{9.0mm}\centering ROU-GE\end{minipage}
 & \begin{minipage}{9.0mm}\centering BLE-URT\end{minipage}
 & \begin{minipage}{9.0mm}\centering LLa-MA\end{minipage}
 & \begin{minipage}{9.0mm}\centering M-P\end{minipage}
 \\
 \midrule

Correct & \R{.492} & \R{.456} & \R{.480} & \R{.802} & \R{.606} \\
+ true & \R{.492} & \R{.456} & \R{.480} & \R{.802} & \R{.606} \\
+ relevant & \R{.488} & \R{.449} & \R{.478} & \R{.808} & \R{.610} \\
+ coherent & \R{.435} & \R{.462} & \R{.466} & \R{.721} & \R{.536} \\

\bottomrule
\end{tabular}}
\resizebox{.33\textwidth}{!}{%
\footnotesize\centering
\setlength{\tabcolsep}{4pt}
\begin{tabular}{l cccccccc}
\toprule
\Large UA: uk
 & \begin{minipage}{9.0mm}\centering chrF\end{minipage}
 & \begin{minipage}{9.0mm}\centering ROU-GE\end{minipage}
 & \begin{minipage}{9.0mm}\centering BLE-URT\end{minipage}
 & \begin{minipage}{9.0mm}\centering LLa-MA\end{minipage}
 & \begin{minipage}{9.0mm}\centering M-P\end{minipage}
 \\
 \midrule

Correct & \R{.399} & \R{.223} & \R{.487} & \R{.805} & \R{.666} \\
+ true & \R{.399} & \R{.223} & \R{.487} & \R{.805} & \R{.666} \\
+ relevant & \R{.399} & \R{.223} & \R{.487} & \R{.805} & \R{.666} \\
+ coherent & \R{.463} & \R{.205} & \R{.509} & \R{.752} & \R{.644} \\

\bottomrule
\end{tabular}}

\resizebox{.33\textwidth}{!}{%
\footnotesize\centering
\setlength{\tabcolsep}{4pt}
\begin{tabular}{l cccccccc}
\toprule
\Large CZ: en
 & \begin{minipage}{9.0mm}\centering chrF\end{minipage}
 & \begin{minipage}{9.0mm}\centering ROU-GE\end{minipage}
 & \begin{minipage}{9.0mm}\centering BLE-URT\end{minipage}
 & \begin{minipage}{9.0mm}\centering LLa-MA\end{minipage}
 & \begin{minipage}{9.0mm}\centering M-P\end{minipage}
 \\
 \midrule

Correct & \R{.392} & \R{.397} & \R{.405} & \R{.760} & \R{.711} \\
+ true & \R{.389} & \R{.396} & \R{.402} & \R{.754} & \R{.700} \\
+ relevant & \R{.381} & \R{.383} & \R{.399} & \R{.693} & \R{.763} \\
+ coherent & \R{.322} & \R{.331} & \R{.331} & \R{.608} & \R{.718} \\

\bottomrule
\end{tabular}}
\resizebox{.33\textwidth}{!}{%
\footnotesize\centering
\setlength{\tabcolsep}{4pt}
\begin{tabular}{l cccccccc}
\toprule
\Large SK: en
 & \begin{minipage}{9.0mm}\centering chrF\end{minipage}
 & \begin{minipage}{9.0mm}\centering ROU-GE\end{minipage}
 & \begin{minipage}{9.0mm}\centering BLE-URT\end{minipage}
 & \begin{minipage}{9.0mm}\centering LLa-MA\end{minipage}
 & \begin{minipage}{9.0mm}\centering M-P\end{minipage}
 \\
 \midrule

Correct & \R{.410} & \R{.428} & \R{.486} & \R{.856} & \R{.786} \\
+ true & \R{.410} & \R{.428} & \R{.486} & \R{.856} & \R{.786} \\
+ relevant & \R{.405} & \R{.423} & \R{.482} & \R{.864} & \R{.793} \\
+ coherent & \R{.379} & \R{.453} & \R{.420} & \R{.730} & \R{.691} \\

\bottomrule
\end{tabular}}
\resizebox{.33\textwidth}{!}{%
\footnotesize\centering
\setlength{\tabcolsep}{4pt}
\begin{tabular}{l cccccccc}
\toprule
\Large UA: en
 & \begin{minipage}{9.0mm}\centering chrF\end{minipage}
 & \begin{minipage}{9.0mm}\centering ROU-GE\end{minipage}
 & \begin{minipage}{9.0mm}\centering BLE-URT\end{minipage}
 & \begin{minipage}{9.0mm}\centering LLa-MA\end{minipage}
 & \begin{minipage}{9.0mm}\centering M-P\end{minipage}
 \\
 \midrule

Correct & \R{.384} & \R{.381} & \R{.479} & \R{.810} & \R{.784} \\
+ true & \R{.384} & \R{.381} & \R{.477} & \R{.813} & \R{.787} \\
+ relevant & \R{.384} & \R{.381} & \R{.477} & \R{.813} & \R{.787} \\
+ coherent & \R{.448} & \R{.426} & \R{.463} & \R{.767} & \R{.730} \\

\bottomrule
\end{tabular}}

\caption{Answer-level point biserial correlation of automatic metrics and human judgment split over countries and languages. Note that each of the tables is only based on a comparison of 5 systems.}\label{tab:answer_level_individual}

\end{table}
\begin{table}

\footnotesize\centering
\begin{tabular}{clcccccc}
\toprule
& \multirow{2}{*}{Model} & \multicolumn{2}{c}{Czechia} & \multicolumn{2}{c}{Slovakia} & \multicolumn{2}{c}{Ukraine} \\ \cmidrule(lr){3-4} \cmidrule(lr){5-6} \cmidrule(lr){7-8}
& & cs & en & sk & en & uk & en \\
\midrule
\multirow{5}{*}{\rotatebox{90}{Textual}} & EuroLLM-9B-Instruct & \confmat{32}{3}{3}{62} & \confmat{25}{7}{11}{57} & \confmat{27}{5}{3}{65} & \confmat{21}{10}{9}{60} & \confmat{24}{3}{2}{71} & \confmat{17}{10}{10}{64} \\[8pt]
 & Llama-3.1-8B-Instruct & \confmat{26}{2}{2}{69} & \confmat{23}{7}{8}{62} & \confmat{17}{2}{1}{80} & \confmat{20}{5}{6}{70} & \confmat{22}{1}{2}{76} & \confmat{22}{8}{7}{63} \\[8pt]
 & Llama-3.3-70B-Instruct & \confmat{55}{3}{2}{39} & \confmat{49}{4}{11}{37} & \confmat{42}{4}{2}{52} & \confmat{41}{4}{7}{47} & \confmat{49}{3}{5}{43} & \confmat{44}{3}{10}{42} \\[8pt]
 & Llama-4-Scout-17B-16E-Instruct & \confmat{37}{3}{2}{59} & \confmat{33}{4}{6}{57} & \confmat{31}{3}{2}{64} & \confmat{28}{5}{6}{61} & \confmat{45}{4}{2}{49} & \confmat{36}{4}{9}{50} \\[8pt]
 & Mistral-7B-Instruct-v0.3 & \confmat{21}{2}{2}{75} & \confmat{20}{5}{5}{71} & \confmat{11}{2}{3}{85} & \confmat{16}{5}{5}{74} & \confmat{19}{3}{2}{76} & \confmat{22}{8}{7}{63} \\
\midrule
\multirow{5}{*}{\rotatebox{90}{Visual}} & gemma3 & \confmat{10}{2}{0}{87} & \confmat{9}{1}{4}{86} & \confmat{14}{5}{0}{81} & \confmat{11}{3}{1}{86} & \confmat{21}{5}{1}{73} & \confmat{18}{1}{3}{77} \\
\bottomrule
\end{tabular}

\vspace{5pt}

\begin{minipage}{0.15\textwidth}
\centering
   Legend: \confmatlegend{} 
\end{minipage}
\begin{minipage}{8cm}
TP/TN color scale:
%

FP/FN color scale:
%
\end{minipage}

\caption{Confusion matrices of LLaMA 3.3 70B Instruct as a judge compared to manually evaluated correctness.}\label{tab:confusion_appendix}

\end{table}

\begin{table}
\centering\footnotesize
\setlength{\tabcolsep}{3.0pt}

\begin{tabular}{l cc cc cc}
    \toprule
    \multirow{2}{*}{Model} & \multicolumn{2}{c}{CZ} & \multicolumn{2}{c}{SK} & \multicolumn{2}{c}{UA} \\ \cmidrule(lr){2-3} \cmidrule(lr){4-5} \cmidrule(lr){6-7}
    & cs & en & sk & en & uk & en \\ \midrule
    \multicolumn{7}{c}{\textit{chrF}} \\ \midrule
    EuroLLM 9B Ins. & 19.9$\pm$3.7 & 19.0$\pm$1.4 & 21.2$\pm$2.0 & 19.7$\pm$1.7 & 20.6$\pm$2.2 & 21.0$\pm$1.5 \\
    Llama 3.1 8B Ins. & 23.3$\pm$0.8 & 20.0$\pm$2.5 & 21.4$\pm$1.2 & 19.4$\pm$2.0 & 25.2$\pm$0.6 & 23.0$\pm$1.5 \\
    Llama 3.3 70B Ins. & 40.2$\pm$2.9 & 30.3$\pm$5.3 & 36.7$\pm$1.3 & 26.2$\pm$4.3 & 38.7$\pm$2.6 & 29.5$\pm$2.7 \\
    Llama 4 Scout Ins. & 28.9$\pm$1.9 & 19.2$\pm$2.2 & 28.4$\pm$1.0 & 19.3$\pm$1.9 & 34.6$\pm$2.8 & 22.9$\pm$3.5 \\
    Mistral 7B Ins. v0.3 & 16.9$\pm$2.0 & 13.4$\pm$1.6 & 17.1$\pm$1.4 & 13.7$\pm$1.9 & 24.5$\pm$1.2 & 16.5$\pm$1.8 \\
    \midrule
    \multicolumn{7}{c}{\textit{LLM as a Judge}} \\ \midrule
    EuroLLM 9B Ins. & 42.8$\pm$1.6 & 35.4$\pm$2.4 & 33.7$\pm$0.7 & 32.5$\pm$0.7 & 27.4$\pm$1.9 & 30.8$\pm$1.5 \\
    Llama 3.1 8B Ins. & 30.5$\pm$1.9 & 31.4$\pm$2.3 & 20.6$\pm$1.3 & 25.4$\pm$3.2 & 25.0$\pm$1.4 & 32.5$\pm$2.2 \\
    Llama 3.3 70B Ins. & 58.0$\pm$1.0 & 58.7$\pm$1.6 & 46.6$\pm$1.4 & 49.5$\pm$1.8 & 53.1$\pm$1.0 & 54.0$\pm$0.6 \\
    Llama 4 Scout Ins. & 41.2$\pm$1.0 & 38.8$\pm$1.3 & 32.5$\pm$1.2 & 33.5$\pm$0.7 & 47.5$\pm$1.5 & 44.3$\pm$2.7 \\
    Mistral 7B Ins. v0.3 & 24.4$\pm$1.3 & 27.4$\pm$0.4 & 17.2$\pm$0.7 & 23.2$\pm$0.7 & 23.4$\pm$1.0 & 32.0$\pm$0.7 \\
    \bottomrule
\end{tabular}

\caption{Average chrF and LLM as a judge scores and their standard deviation for prompt variation.}\label{tab:prompt_variation}

\end{table}

%
%
%
%
%
%
%

\end{document}